\documentclass[sigconf,screen]{acmart}

\usepackage{microtype}
\usepackage{balance} 
\usepackage{booktabs} 
\usepackage{xcolor}
\usepackage{colortbl}
\usepackage{listings, multicol}
\usepackage{multirow} 
\usepackage{enumitem} 
\usepackage{soul}
\usepackage{amsmath}
\usepackage{mathtools}
\usepackage{subcaption}
\usepackage{flushend}
\usepackage[htt]{hyphenat}  
\usepackage[normalem]{ulem}  
\usepackage{array}   
\usepackage{makecell}   
\usepackage[nounderscore]{syntax}
\usepackage{mdframed}
\newcommand{\ftype}{\tau}
\newcommand{\rtype}{\tau_r}
\newcommand{\ttype}{\tau_t}
\newcommand{\rr}{\raggedright}
\newcommand{\tn}{\tabularnewline}
\newcolumntype{L}{>{$}l<{$}} 

\newcommand{\nib}[1]{\noindent{\bf #1}}
\newcommand{\eqspace}{~~~~~}

\newcommand{\sysname}{{{\sc PhysFrame}}}

\newif\iflong
\longtrue 

\setcopyright{rightsretained}
\acmPrice{}
\acmDOI{10.1145/3468264.3468608}
\acmYear{2021}
\copyrightyear{2021}
\acmISBN{978-1-4503-8562-6/21/08}
\acmConference[ESEC/FSE '21]{Proceedings of the 29th ACM Joint European Software Engineering Conference and Symposium on the Foundations of Software Engineering}{August 23--27, 2021}{Athens, Greece}
\acmBooktitle{Proceedings of the 29th ACM Joint European Software Engineering Conference and Symposium on the Foundations of Software Engineering (ESEC/FSE '21), August 23--27, 2021, Athens, Greece}

\begin{document}

\title{\sysname: Type Checking Physical Frames of Reference for Robotic Systems}
\iflong
\thanks{This is an extended version of an article published in Proc. ESEC/FSE 2021 [12 pages, \url{https://doi.org/10.1145/3468264.3468608}].}
\else
\thanks{An extended version is available at arXiv~\cite{FullVersion}.}
\fi

\author{Sayali Kate}
\affiliation{%
  \institution{Purdue University}
  \country{USA}
}
\email{skate@purdue.edu}

\author{Michael Chinn}
\affiliation{%
  \institution{University of Virginia}
  \country{USA}
}
\email{mec2wr@virginia.edu}

\author{Hongjun Choi}
\affiliation{%
  \institution{Purdue University}
  \country{USA}
}
\email{choi293@purdue.edu}

\author{Xiangyu Zhang}
\affiliation{%
 \institution{Purdue University}
 \country{USA}
}
\email{xyzhang@cs.purdue.edu}

\author{Sebastian Elbaum}
\affiliation{%
  \institution{University of Virginia}
  \country{USA}
}
\email{selbaum@virginia.edu}

\renewcommand{\shortauthors}{S. Kate, M. Chinn, H. Choi, X. Zhang, S. Elbaum}

\begin{abstract}
A robotic system continuously measures its own motions and the external world during operation. Such measurements are with respect to some frame of reference, i.e., a coordinate system. A nontrivial robotic system has a large number of different frames and data have to be translated back-and-forth from a frame to another. The onus is on the developers to get such translation right. However, this is very challenging and error-prone, evidenced by the large number of questions and issues related to frame uses on developers' forum. Since any state variable can be associated with some frame, reference frames can be naturally modeled as variable types. We hence develop a novel type system that can automatically infer variables' frame types and in turn detect any type inconsistencies and violations of frame conventions. The evaluation on a set of 180  publicly available ROS projects shows that our system can detect 190 inconsistencies with 154 true positives. We reported 52 to developers and received 18 responses so far, with 15 fixed/acknowledged. Our technique also finds 45 violations of common practices.
\end{abstract}

\ccsdesc[500]{Software and its engineering~Abstract data types}
\ccsdesc[500]{Software and its engineering~Software defect analysis}

\keywords{Physical Frame of Reference, Frame Consistency, Type Checking, Static Analysis, z-score Mining, Robotic Systems}

\maketitle

\section{Introduction}
\label{sec:intro}

Robotic systems have rapidly growing applications in our daily life, enabled by the advances in many areas such as AI. Engineering such systems becomes increasingly important.
Due to the unique characteristics of such systems, e.g., the need of modeling the physical world and satisfying the real time and resource constraints, robotic system engineering poses new challenges to developers. One of the prominent challenges is to properly use physical frames of reference.
Specifically, the operation of a robotic system involves moving individual body parts and interacting with the external world.
It entails precisely measuring positions and orientations of body parts and external objects. All these measurements are represented with respect to a set of coordinate systems (also called {\em frames of reference} or {\em frames} in short). For example, in a three-dimensional coordinate system, the location (1,2,3) is meaningless unless we know the frame to which this location refers to. The origin of a frame provides the reference position (0,0,0), whereas the orientation of the three axes tells us the directions in which they point to. Further, the origin and orientation of a frame itself may be defined relative to another frame and so on.

Robotic systems usually follow a
modular design, in which body
parts and control software components are developed independently by various parties. As such, different components
often use different frames. 
For example, a camera has the {\em camera frame} through which the physical world is measured from the camera's perspective. In this frame, the center of camera is the origin and the axes follow the orientations of the camera.  
The body of a robot has the {\em body frame} whose origin is the center of the body and axes are pre-defined based on the shape of robot. 
Measurements in the camera frame have to be translated to the body frame before they can be used in body related computation.
Since the camera can be mounted at different places and moves during operation, such translation has to be done by the developers in software.
More discussion about frames and an example can be found in Section~\ref{sec:background}.

\begin{figure}[!ht]
 \setlength{\abovecaptionskip}{1pt}
    \centering
    \includegraphics[width=\columnwidth]{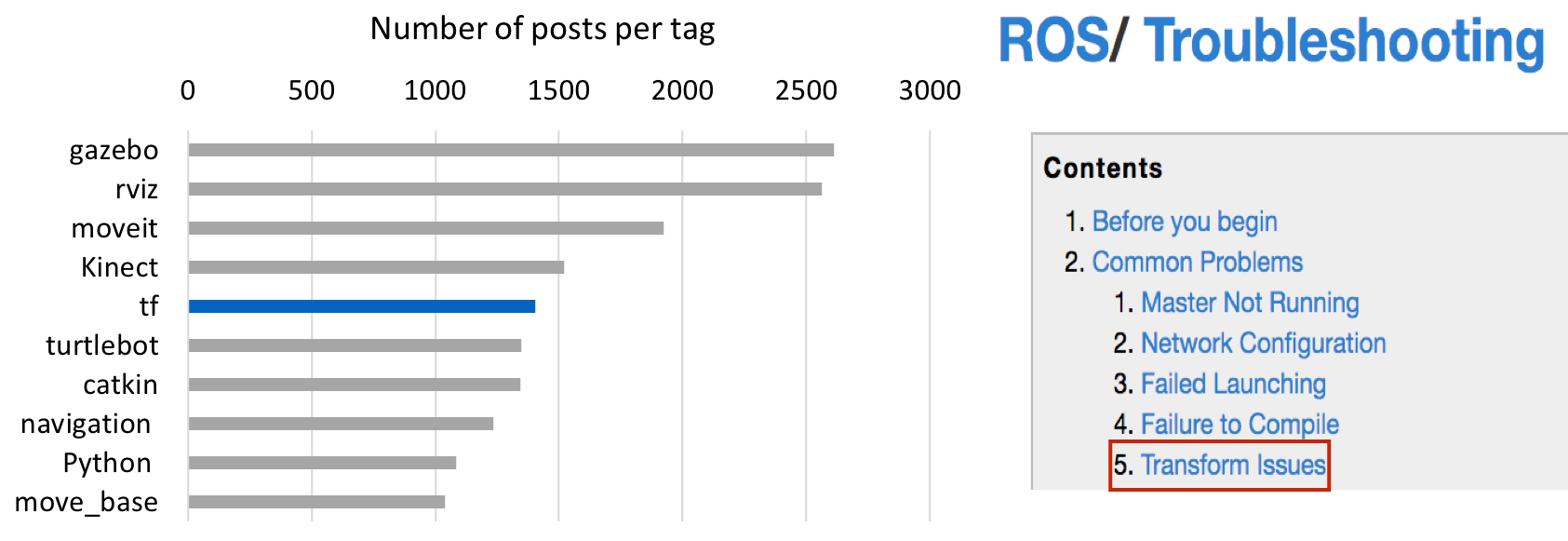}
    \vspace{-0.15in}
    \caption{Forum posts that show frames in robotic programming are difficult to handle. {\footnotesize (Note: right-side snapshot is taken from http://wiki.ros.org/ROS/Troubleshooting page licensed under the CC BY 3.0)}}
    \vspace{-0.20in}
    \label{fig:posts}
\end{figure}

Most robotic systems are developed in general purpose languages such as C/C++, facilitated by domain specific libraries. 
Such languages do not have intrinsic support for the additional complexity induced by the use of frames. For example, although popular libraries such as ROS~\cite{quigley2009ros} provide functions to facilitate translation between frames, the onus is on the developers to determine
the reference frames of program variables, the places where translation is needed, and correctly implement the concrete translations. Since translation is often done by vector/matrix operations, many developers even realize their own translation functions from scratch without using ROS APIs. 
As a result, use of frames is error-prone, even for experienced developers.  
This is evidenced by the fact 
that a large number of questions and issues for a robotic system project are usually regarding reference frames. 
Figure~\ref{fig:posts} shows that questions tagged with `{\it tf}' (a keyword for ROS frame support
\cite{6556373,rostfpkg}
) are among the top ten types of questions by ROS developers, such as``{\it confused about coordinate frames}", ``{\it \emph{tf} transforms are confusing}", and ``{\it is there an easier way to find the coordinates of a point in another frame}". Also, ROS wiki identifies transform issues as one of the top 5 common problems. 
Besides the difficulties of getting them right during development, 
misuse of frames may cause robot runtime malfunction~\cite{frameissue1}, code reuse problems~\cite{frameissue2,frameissue4}, and maintenance difficulties after deployment.

ROS provides a number of tools to help developers debug frame related problems~\cite{tftools}.  
These are mostly runtime tools that can facilitate visualization of frames, relation between frames, details of a transformation between two frames (e.g. when it is created and by which component). 
However, it is always more desirable to detect problems as early as possible in the development life-cycle.
A static tool that can scan and detect frame misuses before running the system would be highly desirable. 
To the best of our knowledge, there are unfortunately no such tools.

In this paper, we introduce and implement a novel fully automated static tool to identify frame related problems in ROS-based projects. Developers direct the tool to operate on a  project directory, and the tool automatically models the project, and checks for potential frame faults.
Since  each physical state related variable (e.g., sensor reading, acceleration, velocity, and position) is associated with some frame, our tool builds on such such associations which are similar to variable types by nature. We hence propose a type system to model reference frames.
A traditional type system often starts with type annotations that require developers to know variable types in the first place, which is difficult in our context as knowing the right frames is usually difficult.   
Instead, our technique automatically infers frames and performs type consistency checks,
leveraging ROS conventions  that can be extracted from its specifications and mined from code repositories. 
The technique uses frame information sources such as static data values exchanged between components and the ROS frame APIs for assigning frame types, and defines type propagation and checking rules based on the standard coordinate conventions documented in ROS-Enhancement-Proposals (REPs) and the semantics of frame related operations such as displacements and rotations. 
As part of our contribution we  present a type language as a vehicle to describe these type rules , which constitute the foundation for our domain-specific static analyzer.

Our contributions are summarized as follows:
\begin{itemize}[leftmargin=*]
    \item We propose a 
    fully automated type inference and checking technique for physical frames in ROS-based programs to detect frame inconsistencies and convention violations.
    \item We define frame and transformation abstract types, and our system automatically infers such types for variables.
    This is achieved by novel abstraction of frame related program semantics.
    \item We present a data-driven approach to identifying commonly followed practices that may not be documented anywhere. Our system also checks for violations of such practices.
    \item We implement a tool, \sysname{}, and evaluate it on 180 ROS projects from Github. It 
    reports 190 type inconsistencies with 154 true positives (81.05\%). We report 52 from projects that are recently active to developers and have received 18 responses so far, with 15 fixed/confirmed.
    It also detects 45 violations of common practices.
\end{itemize}

\section{Reference Frames and ROS}
\label{sec:background}

\subsection{Frames}

A coordinate system that is used as a reference for measurements of quantities such as position and velocity is called a {\em frame of reference}, in short, a frame. It is defined by two components: an origin point and the axes system. 
A robotic system often consists of many parts, each having at least a {\em body frame} with the origin at the center of the rigid shape of the part, and axes pointing to some orthogonal directions.
It allows measurements in the perspective of the part.
The body frame for the base of a robot is often called {\em base\_link}. It is one of the most important frames in a robot as measurements in this frame describe how the robot as a whole sees the world. The body frame for a camera is called the {\em camera frame}.
If the part is a sensor, it often has other frames to denote sensor readings. For example, a camera has multiple {\em optical frames}, including the {\em color frame} measuring RGB format image, the {\em depth frame} measuring depth image, and the \emph{ir frame} measuring infra red image.
These frames are centered at the corresponding sensors (e.g., lens) instead of the camera body.
They are different from the camera frame. 

There are also the {\em odometry (odom) frame}, the {\em map frame}, and the {\em earth frame} to denote the physical world, i.e., how the world sees the robot and its surroundings. The {\it odom} frame represents robot's initial pose and does not move with the robot. It is like a third person standing at the robot's starting position and observing the motion of robot in the environment.
The {\it map} frame projects everything in a localized map such as inside a room.
It is like a third person standing somewhere in the room 
(not the starting point of the robot) and observing the robot and its environment. 
The {\it earth} frame is an earth centered frame. 

During operation, sensor readings are acquired regarding the corresponding sensor frames. They are then transformed to the body frames of the parts where the sensors are installed, and eventually to the {\it base\_link} frame and some world frame(s) in which control algorithms use the transformed values to compute actuation signals. These signals may undergo the inverse transformations until they reach the actuators.
A simple commodity robot may have up to 81 different frames~\cite{pr2} as shown in Figure 7
in Appendix
\iflong \else in~\cite{FullVersion} \fi
and values have to be correctly transformed between any pair of frames. 
It is a heavy burden for developers to get all these transformations right.

\begin{figure}
    \centering
    \includegraphics[width=0.95\columnwidth]{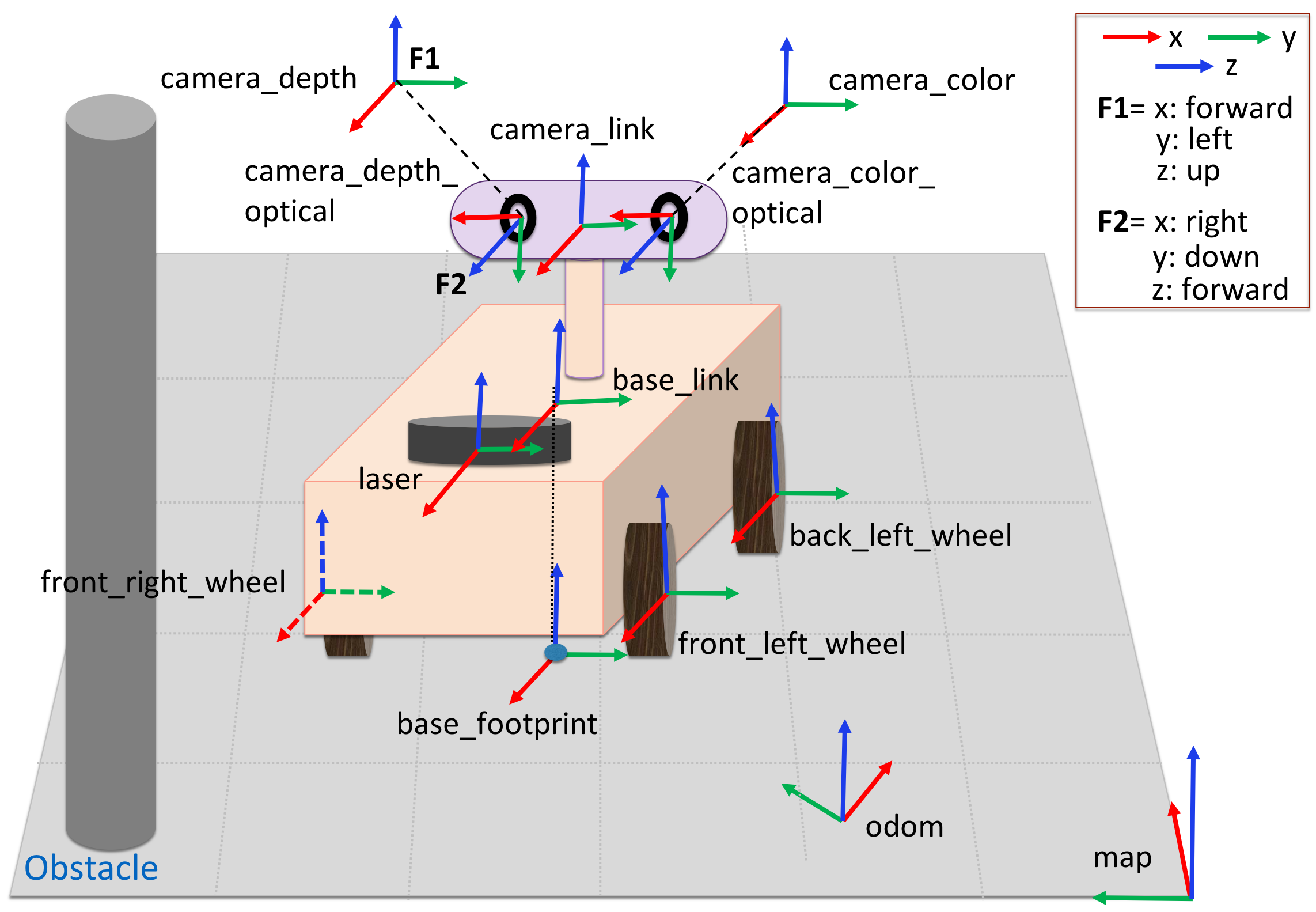}
    \vspace{-0.10in}
    \caption{Frames in a Simple Robot}
    \vspace{-0.23in}
    \label{fig:frames}
\end{figure}

Figure~\ref{fig:frames} shows a simple robot. Observe that each of its parts has its own body frame (e.g., {\it camera\_link}, {\it laser}, and {\it base\_link}).
There are also the world frames such as {\it odom} and {\it map}.
Suppose if the vehicle locates an obstacle, then it knows the obstacle's position in the \emph{camera\_depth\_optical} frame.
This position is different from the position of the obstacle in the \emph{map} frame. In order to answer the question 
``{what is the position of an obstacle in the room?}", the vehicle needs to know 
how the \emph{camera\_depth\_optical} frame is positioned and oriented relative to the \emph{base\_link} frame and 
how the \emph{base\_link} frame is positioned and oriented relative to the \emph{map} frame,
then has to convert the obstacle position from the \emph{camera\_depth\_optical} frame 
all the way to
the \emph{map} frame.
Moreover, the position of the obstacle in the frame of front right wheel, \emph{front\_right\_wheel}, is different from that in the \emph{base\_link} frame. Hence, to know the position with respect to the front right wheel, the vehicle needs to know the position and orientation of \emph{front\_right\_wheel} frame relative to the \emph{base\_link} frame.
Different body frames may have different axis orientation conventions.
Figure~\ref{fig:frames} presents two axis orientations labeled as F1 and F2. Each part has a default forward-facing direction (e.g., the nose of an aerial robot and the front of a camera), the values {\it forward, backward, left, right}, etc., are defined regarding the default direction. F1 and F2 are widely used in ROS. Moreover, various libraries use different conventions. For example, (x: right, y: up, z: backward) axis orientation is widely used in OpenGL.

\subsection{ROS and Frames}
\label{subsec:ros-conv}
The Robotic Operating System (ROS)~\cite{quigley2009ros} is an open source software framework for building robotic applications. Its modular, distributed nature has made it widely used. Its users range from a novice developer working on a hobby project to industries. ROS provides various components such as drivers, localization packages, odometer packages developed by the community in addition to the core components. 
It provides a package called \emph{tf} or \emph{tf2}~\cite{6556373,rostfpkg} to support transformation between frames. 

\smallskip
\noindent
{\bf ROS Transforms.}
To specify frames and transformations, developers leverage the {\em tf} library to explicitly publish 
transformations  between frames (called {\em transforms} in ROS), which also implicitly introduce the frames.
A transform consists of a parent frame and a child frame, uniquely identified by their ids. ROS developers often say a transform is from the parent to the child. For example ``{\em map}$\rightarrow${\em odom}'' means a transform from the {\it map} frame to the {\em odom} frame. It is defined by providing the displacements of the child frame's origin and the rotation of its axes regarding the parent frame. However, {\em the transform is used to translate states in the child frame to the parent frame} (not the other way suggested by its name). We  unfortunately have to inherit such term ambiguity from ROS. In the rest of the paper, we explicitly distinguish the name of a transform and its use in state translation to avoid confusion.  

After transforms are published, other ROS components, such as standard control algorithms, subscribe to these transforms based on their names and use them in computation without knowing how they are implemented.  Hence, these core algorithms can be agnostic to concrete system composition.

\noindent
{\bf Launch Files.}
Many parts have fixed and static relative positions, e.g., camera mounted at a fixed position of an aerial robot. 
The transforms between their frames can be statically 
defined in {\em launch files}, which are a kind of static initialization files that define how to start
executing a robotic system. The transforms for frames with dynamic relative positions, e.g., when a camera 
is mounted on a gimbal, have to be published and updated on-the-fly during operation. They are hence implemented by code.
An example launch file snippet can be found in Appendix A
\iflong \else in~\cite{FullVersion}\fi.

\noindent
{\bf TF Tree.}
Transforms are potentially needed in between each pair of frames. However, specifying/implementing transforms for each pair entails substantial and error-prone efforts. ROS has a clean hierarchical design to reduce the complexity. If we consider a
frame as a node and a transform between two frames as an edge, ROS requires all published frames and transforms
to form a tree, that is, no cycles and each node having only one parent. We call it a {\em TF tree}. As such, the developer only needs to 
focus on realizing the transforms denoting individual edges. ROS then automatically constructs the transform
from an arbitrary frame $f_1$ to another arbitrary frame $f_2$, by first performing the transforms from $f_1$ to the lowest common ancestor frame of the two, and then the transform from the ancestor to $f_2$. It can be easily inferred that such a design substantially reduces the developers' burden. 
The below figure shows the TF tree for the robot in Figure~\ref{fig:frames}. 
Observe that while the developers only need to develop and publish 12 transforms (denoted by the edges), ROS allows 156 transforms from a frame to any other frame. 
The red edges form a transform path between \emph{front\_right\_wheel} to \emph{camera\_depth\_optical}, when the right wheel on the front of the robot uses data from the camera depth sensor.
\vspace{-0.05in}
\begin{center}
    \includegraphics[width=\columnwidth]{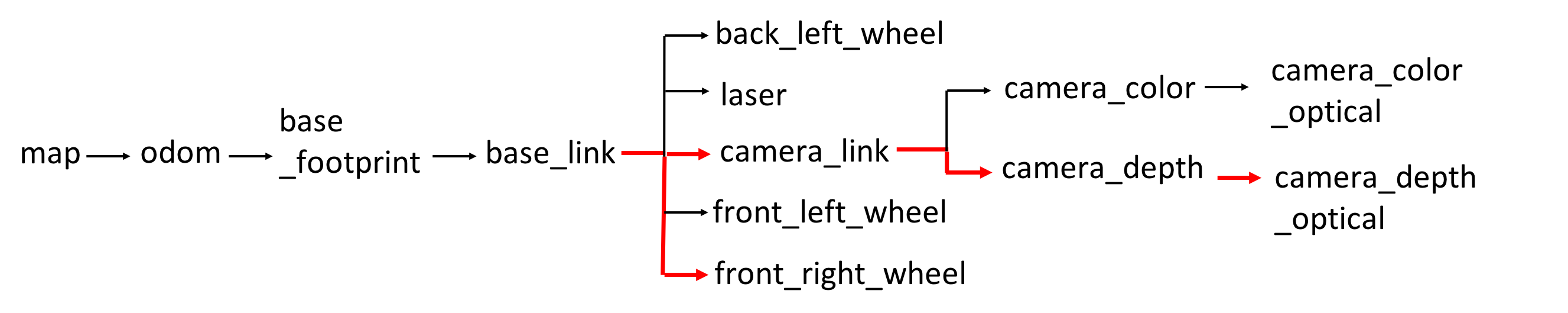}
\end{center}
\vspace{-0.15in}

\subsection{Frame Conventions in ROS}
\label{subsec:conventions}
All ROS projects follow certain conventions when manipulating frames. Violation to these conventions would cause integration, maintenance, and reuse problems. 
Our technique leverages these conventions to automatically 
infer and check frame types. 
There are two sources for extracting conventions: ROS specification and launch files.
The former is explicit and the latter is implicit.
Since ROS conventions are often related to static coefficients values (e.g., orthogonal axes orientations), they are difficult to extract directly from code where coefficients are largely variables. 

\noindent
{\bf Explicit Conventions From ROS Specifications.}
These conventions can be classified into 
the following three categories.

\noindent
\underline{\em Naming Conventions.}
ROS applications use messages to exchange data between components. The {\tt frame\_id} field in a message specifies the frame for the data in that message. 
These ids follow certain conventions. 
For example, ``{\tt map}'' is used for the map frame~\cite{rep105}, and ``{\tt base\_link}'' for the body frame of the robot's base~\cite{rep105,rep120}; optical frames should have an ``{\tt \_optical}'' suffix~\cite{rep103}.

\noindent
\underline{\em Axis Orientation Conventions.}
According to ~\cite{rep103},
body frames should have the (x: forward, y: left, z: up) or FLU axis orientation 
(i.e., F1 orientation in Figure~\ref{fig:frames}); 
optical frames should have the (x: right, y: down, z: forward) orientation  
(i.e., F2 in Figure~\ref{fig:frames}).

\noindent
\underline{\em Tree Order Conventions.}
Recall that in a robotic system, the published frames and transforms are organized in a TF tree. ROS requires TF trees follow a partial order:  {\it earth} $\rightarrow$ {\it map} $\rightarrow$ {\it odom} $\rightarrow$ {\it base\_link}~\cite{rep105}.
Note that while a concrete TF tree for a specific robotic system may omit some of these frames and have additional frames, such an order must be respected. The reason is that the order optimizes the performance for
mostly commonly seen transforms. 

\noindent
{\bf Implicit Conventions from ROS Launch Files.}
Launch files provide a rich resource for mining implicit conventions. 
We focus on extracting two kinds of implicit conventions.

\noindent
\underline{\em Co-occurrence Conventions.} Robotic systems of a same kind share similarity in their physical compositions. As such, they
often use similar frames and transforms. This is reflected by co-occurrences of transforms in launch files. For example, a ground vehicle robot with two wheels has one wheel on its left and the other wheel on its right. If a transform from \emph{base\_link} to the frame of left wheel is defined, then a transform from \emph{base\_link} to the frame of right wheel is also defined. 

\noindent
\underline{\em Value Conventions.}
Certain transforms are associated with fixed coefficient values such as 0 rotation angles or 0 displacement values. For example, a transform from \emph{camera\_depth} to \emph{camera\_depth\_optical} must have 0 displacement values.

\vspace{-0.10in}
\section{Motivation}
\label{sec:motivation}

\begin{figure}[!htbp]
    \centering
     \includegraphics[width=\columnwidth]{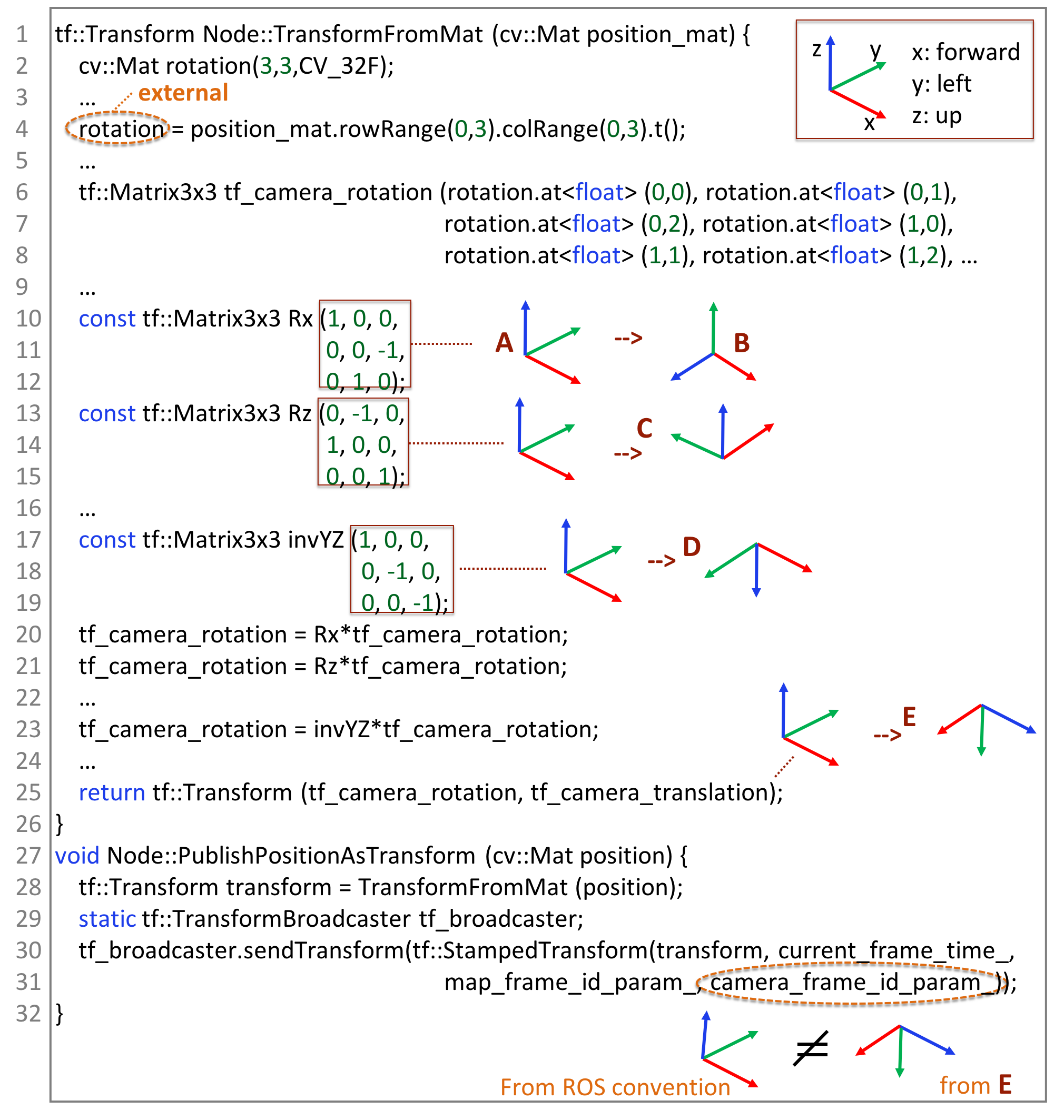}
     \vspace{-0.20in}
    \caption{Example of incorrect transform between frames \newline {\footnotesize source: https://git.io/JtSHb, fixed source: https://git.io/JtSRt}
    }
    \vspace{-0.20in}
    \label{fig:motiv}
\end{figure}

To motivate our technique we use code from the {\em simultaneous localization and mapping} (SLAM) component  ORB2-SLAM2 (links in the caption of Figure~\ref{fig:motiv}) \cite{orb-slam2-ros,murORB2}. 
Localization and mapping algorithms like ORB are key to the operation of modern mobile robots, incrementally building a map of the robot surroundings utilizing  its {\it camera} as the robot navigates. 
This component, however, uses different conventions from ROS and hence the most important task for interfacing then is to perform appropriate frame conversions. 

The code snippet in Figure~\ref{fig:motiv} aims to implement a transform 
to facilitate translation of states from the ROS {\em camera} frame to the ROS {\em map} frame using the environment model of ORB-SLAM2. The conversion implementation, however,
is incorrect, and \sysname{} catches this error by type checking.

Constructing the aforementioned transform is non-trivial, and consists of three groups of operations.
First, given some ROS state in the child {\it camera} frame with the conventional FLU orientation, it should be translated into ORB's RDF orientation,
({x:right}, { y:down}, { z:forward}).
Second,
the position of the {\em camera} frame in the {\em map} frame needs to be determined.
However, since ORB-SLAM2 models the environment in the {\em camera} frame, one can only query the position of {\em map} frame in the {\em camera} frame, so it needs to be transposed to acquire 
the position of {\em camera} frame in the {\em map} frame.
Third,  the transform should  
translate the results back to ROS's FLU orientation. The  code shown  fails to accomplish the last step.

Function {\tt PublishPositionAsTransform()} starting at line 27 takes the ORB-SLAM2's position matrix (denoting the {\it map} frame's position in the {\it camera} frame), i.e., variable {\tt position}, and 
constructs a ROS transform (line 28) by invoking {\tt TransformFromMat()}, which starts at line 1.
In line 4 it acquires the rotation matrix from the ORB-SLAM2's position matrix,
which includes both displacements and rotation, and performs transposition by function {\tt t()}.
Line 6 clones it to {\tt tf\_camera\_rotation}.
Line 10 creates a matrix denoting rotation of an FLU axis system anti clockwise for 90 degree with respect to the $x$ axis (shown to the right of the code). Lines 13 and 17 create two additional rotation matrices.
Lines 20-23 compose the three matrices and the earlier rotation matrix.
Line 25 creates a ROS transform from displacements denoted by {\tt tf\_camera\_translation} and rotation denoted by {\tt tf\_camera\_rotation}. When the transform is used in state translation, a state is multiplied with {\tt tf\_camera\_rotation}, which equals to {\tt invYZ*Rz*Rx*rotation}. The first three matrixes are equivalent to transforming the FLU orientation of the state to the RDF orientation as shown on the right of line 25. The fourth matrix ({\tt rotation}) then translates the state from the camera frame
to the map frame.
However, the constructed transform is problematic as it forgets to further translate back to FLU.
This bug will cause downstream system mis-behaviors, which could be devastating if the camera readings
are the dominant source for control decisions. It is fixed by appending the inverse rotation matrices of {\tt invYZ}, {\tt Rz}, and {\tt Rx}  to {\tt tf\_camera\_rotation} after line 23, to change RDF to FLU.

\sysname{}  associates each state variable with a frame type that abstracts both the displacements and orientation of the frame in its parent frame. It also associates each transform variable with a transform type that denotes the displacements and rotation needed in frame translation. It propagates such types following operation semantics and checks consistency. In the example code, 
it associates matrix {\tt Rx} at line 10 with a transform type from the FLU orientation $A$ to the orientation $B$. The type is derived from the constant matrix values. 
It further associates matrix {\tt Rz} at line 13 with a transform type from FLU to the orientation $C$; and {\tt invYZ} at line 17 from FLU to $D$.
By modeling the semantics of matrix multiplications at lines 20, 21, and 23, it determines that the transform at line 25 has the type 
from FLU to RDF. 
This yields a type error at
lines 30-31 when the transform is published by function {\tt sendTransform()} because ROS convention demands both  camera frame and map frame to have FLU, whereas the transform converts FLU to RDF. 

Observe that it is difficult to get such transformation right due to its intrinsic subtlety and complexity. 
While this is only a one-step transform, as mentioned earlier, a non-trivial robotic system has more than 80 different frames and transforms between any pair may be necessary. 
The example bug is just one of the many kinds of bugs \sysname{} detects. Others include missing frame and broken TF tree. More discussion can be found in Section~\ref{sec:typeerrors}.


\section{Our Approach}
\label{sec:approach}

\subsection{Overview}
\label{subsec:overview}

\begin{figure}[!t]
    \centering
    \includegraphics[width=\columnwidth]{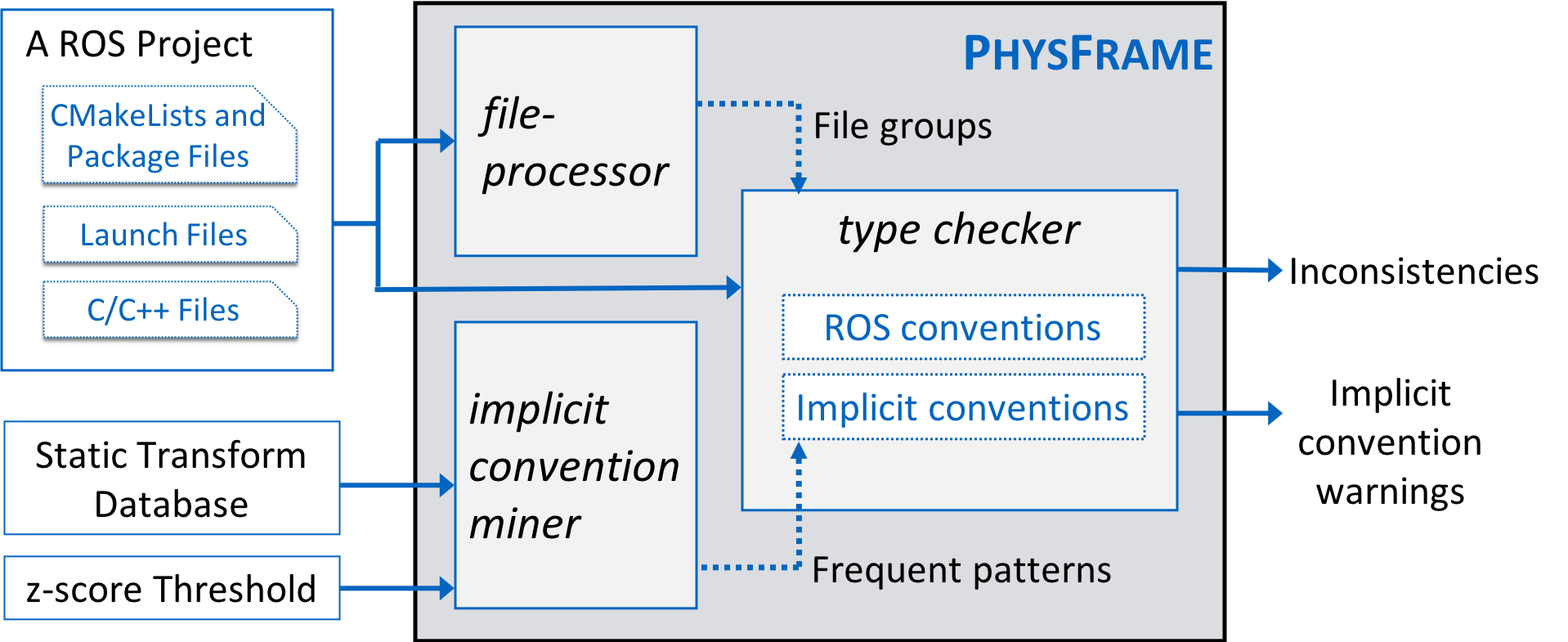}
    \vspace{-0.15in}
    \caption{Overview of \sysname}
    \vspace{-0.20in}
    \label{fig:overview}
\end{figure}

Figure~\ref{fig:overview} presents an overview of \sysname. It takes as input a C++ ROS project, and outputs frame inconsistencies and implicit convention violations. 
\sysname{} consists of three components: {\em implicit convention miner}, {\em file-processor}, and {\em type checker}.

The miner takes a database of static transforms extracted from the launch files of a large repository of over 2200  
mature ROS projects from Github and performs data mining to infer implicit conventions. The miner  identifies frequent static patterns that include pair of transforms that commonly co-exist in a project, transforms commonly having zero displacements, and transforms commonly having zero rotation. Since these patterns may happen coincidentally, the miner computes a frequency  for each pattern to quantify its certainty (we use z-score~\cite{z-score} as a frequency measure) and reports it if it is above a specified z-score threshold. The mining process is executed once on the collected static transform database, but can be updated as new static transforms are incorporated into the database. The produced rule conventions are then used in the analysis for each subject ROS project to identify anti-patterns.
Details can be found in Appendix B
\iflong \else in~\cite{FullVersion} \fi.

A ROS project 
often contains multiple executable subsystems, each having its own TF-tree. 
\sysname{} has to analyze these subsystems one at a time. The file-processor separates a project into subsystems, each consisting of a group of C/C++ source files and launch files that are executed together.
A subsystem usually has a separate initial launch file. 
The file-processor hence starts from these top level launch files, identifies the child launch files and the corresponding CMakeLists files, which indicate the source files involved in each subsystem, and in turn finds file groups that represent the project's subsystems.
Details are elided. 

The type checker then takes patterns and file groups information, together with project files, and types variables and statements in the C/C++ source files.
Type errors are reported when the program cannot be properly typed. 
\sysname{} reports 7 kinds of type errors and 
2 kinds of type warnings as inconsistencies, and reports 3 kinds of implicit convention violation warnings (see Section~\ref{sec:typeerrors}).
Warnings may not lead to broken functionalities but problems in code maintenance and reuse.
In the following, we will focus on explaining the type system.

\vspace{-0.10in}
\subsection{Type System}
\label{subsec:typesys}

Before introducing the formal language and type rules, we intuitively explain how frames and transforms are used.
Typically, one-step transform objects are explicitly created to facilitate translation of states in a child frame (e.g., position) to a parent frame.
A transform specifies the relative position and orientations of the child frame in the parent frame, denoted by displacements and rotations.
It also specifies the (string) ids for the child and parent frames. An id is the unique representation of a frame. 
The creations of transforms also implicitly define frames (through the ids). They are implicit as there are no explicit ROS data structures for frames. 
A state variable can be explicitly associated  with some 
frame id, yielding a {\em stamped}
(a ROS term)
state variable.
A transform can be explicitly applied to some stamped variable (in a child frame) to acquire its correspondence in the parent frame.
One-step transforms can be published and hence become part of the global TF tree. As such, transforms between arbitrary frames (e.g., those multiple steps apart) can be automatically constructed by traversing the tree (Section~\ref{subsec:ros-conv}).
To define a one-step transform, developers often explicitly specify the entailed rotation. A rotation can be composed from others. For example, a 60 degree rotation can be composed from two 30 degree rotations.
Plain state variables (not stamped) can be operated on just like regular C/C++ variables/data-structures. 
As such, even they (implicitly) belong to some frames, frame inconsistencies cannot be detected. Furthermore, ROS does not provide type checking mechanism even for stamped variables. The C/C++ type system cannot offer help either as it does not have any domain knowledge. For example, it can type a variable to a ROS transform data structure but cannot identify the specific transform type, i.e., the parent and child frames and the displacements and rotation involved.

\subsubsection{\bf A Simplified Language and Types.}
\label{subsec:types}
ROS is based on C/C++ with a set of library functions. While our type system supports the complex syntax of C/C++/ROS and models the frame related APIs, we use a simplified language for discussion brevity.
The language focuses on modeling frame related semantics and ignores the standard C/C++ types, operations and statements.
It directly models a number of ROS {\em tf} APIs as language primitives.
Moreover, since our analysis is flow insensitive, control structure like conditional statements and loops are less relevant and hence not modeled. 

\noindent
{\bf Types.}
We define three types: (1) frame type ($\mathit{Frame}$) that represents the frame information of a state variable, (2)  transform type ($\mathit{Transform}$) that represents frame transforms, and (3) rotation type ($Rotation$) that represents changes in the axes' directions. Note that these are not ROS types, but rather types in \sysname{}. 

\noindent
\underline{\em Frame Type.} 
This is an abstract type and always used with some concrete state type such as {\em Point} and {\em Pose}, indicating points  and poses in a specific frame, respectively. 
For example, ROS uses {\tt tf::Stamped<T>(T x, ..., String id)} to denote a {\tt T} type state object associated with a frame identified by {\tt id}. 
The frame type's definition is as follows.
\begin{equation}
\setlength{\abovedisplayshortskip}{0.5pt}
    Frame::= \eqspace <id, pid, x, y, z, d_x, d_y, d_z> 
\end{equation}
It is composed of eight fields:  $id$ denoting the (string) name of frame, $pid$ the name of its parent frame type;
the next three fields, $x,y,z$, being the positions of the frame's origin with respect to its parent frame, where their values are of type \textit{Displacement}.
\begin{equation}
\setlength{\abovedisplayshortskip}{0.5pt}
    \textit{Displacement}::= R \eqspace|\eqspace \top
\end{equation}
where, $R$ is a static numerical value, and $\top$ represents a value unknown statically.
The next three fields, $d_x,d_y,d_z$, encode the $x$, $y$, $z$ axes orientations of a frame, with values of the
\textit{Orientation} type:
\vspace{-0.05in}
$$\mathit{Orientation}=\{\textit{left, right, up, down, forward, backward}\}$$
They denote the (orthogonal) orientations of
axes {\em regarding the body part corresponding to the frame}. They are independent of the placement of the body part regarding its base. For example, although a camera may be installed with arbitrary (and even dynamic) angles regarding a robot's body, the orientation of the camera frame's axes is {\em forward-left-up} (FLU), which is orthogonal regarding the camera.
Although axes orientations are orthogonal, different components and robotic frameworks have different orientation conventions, transformations need to be explicitly performed by developers. 

\noindent
\underline{\em Transform Type.} ROS allows developers to create transform objects that denote transformation from a frame to another (e.g., using the {\tt tt::Transformer} class). We associate such objects with a transform type that abstracts information to facilitate consistency and convention checks. The transform type is defined as follows.  
\begin{equation}
\setlength{\abovedisplayshortskip}{0.5pt}
\setlength{\belowdisplayskip}{0.5pt}
    \begin{split}
        Transform::= <\mathit{cid}, \mathit{pid}, x, y, z, \rtype> 
    \end{split}
\end{equation}
Here, $cid$ and $pid$ are the child and parent frame ids, respectively, $x$, $y$, $z$ are abstract values of the aforementioned {\it Displacement} type, and $\rtype$ denotes an abstract rotation, which will be explained next.
The displacements and rotation together define a
transformation.

\noindent
\underline{\em Rotation Type.}
Rotations change axes orientations of a frame. 
ROS allows creation of  first-order rotation objects (e.g., through the ROS {\tt tt::Transform} class)
and manipulations of these objects. They can be further used to construct transform objects. In this work, we are interested in orthogonal rotations such as those with 90 or 180 degrees as they are used in changing axes orientations. We hence define rotation type as follows.
\begin{equation}
\setlength{\abovedisplayshortskip}{0.5pt}
\setlength{\belowdisplayskip}{0.5pt}
    \begin{split}
        Rotation::= \eqspace <ds_x, ds_y, ds_z> \ | \ \top
    \end{split}
\end{equation}
The three fields, $ds_x, ds_y, ds_z$ are variables of {\it DirSwitch} type:
\vspace{-0.05in}
$$\mathit{DirSwitch} = \{``x",\ ``-x",\ ``y",\ ``-y",\ ``z",\ ``-z",\ \top\}$$

Field $ds_x$ holds a value indicating that
the axis denoted by the value points in the same direction as the previous $x$ axis (i.e., the $x$ axis before rotation). Fields $ds_y$ and $ds_z$ work in a similar way.
We use the following example to explain the semantics.
Let $<``z",\ ``-x",\ ``-y">$ be a {\it Rotation} type. It switches  orientations of axes such that $z$ axis points in the same direction as the previous $x$ axis' direction, $-x$ axis points in the same direction as the previous $y$ axis' direction (i.e., the current $x$ axis is the opposite of the previous $y$ axis), and $-y$ points in the same direction as the previous $z$ axis' direction. 
Assume a variable of 
frame type $<$\textit{ ..., forward, left, up}$>$. After applying the aforementioned rotation, a new frame type 
$<\textit{..., right, down, forward}>$ is derived. Value $\top$ means the rotation is not orthogonal or cannot be determined statically.

\begin{figure}[t]
    \begin{mdframed}
      {\footnotesize \tt
  \begin{grammar}
    <$Program$> $P$ $\Coloneqq$ $S$
    
    <$Statement$> $S$ $\Coloneqq$ $S_1$;$S_2$ | $x:=v$ | $x:=y$ | $x:=y$ \textbf{op} $z$ | \\ 
    $\mathit{fx}:=$ \textbf{stamped}$(s,x)$ |
    $\mathit{fx.id}:= s$ | \\
    $x:=$ \textbf{get\_data} $(\mathit{fx})$ | 
    $\mathit{fx}:=\mathit{fy}$ |  $\mathit{fx}:=$\textbf{external}$()$ | \\
    $\mathit{fx}:=$ \textbf{transform\_to}$(s,\mathit{fy})$ | \\
    $\mathit{tx} :=$ \textbf{new\_transform}$(s_{chd},s_{pnt}, x,y,z,\mathit{rx})$ | \\
    \textbf{sendTransform} $(\mathit{tx})$ | \\
    $\mathit{tx} :=$ \textbf{lookupTransform} $(s_{pnt},s_{chd})$ | \\
    $\mathit{fx}:=$ \textbf{apply\_transform} $(\mathit{tx},\mathit{fy})$ | $\mathit{rx}:= v_{3\times 3}$ | \\
      $\mathit{rx}:=x$ | $\mathit{rx}:= \mathit{ry}\otimes \mathit{rz}$ |
    \textbf{publish} $(s_{topic}, \mathit{fx})$ 
  \end{grammar}    
  \begin{tabular}{ll} 
    <$StateVar$>&  $x,y,z$\\
    
    <$StampedStateVar$> & $\mathit{fx}, \mathit{fy}, \mathit{fz}$
\\    
    <$TransformVar$> & $\mathit{tx}, \mathit{ty}, \mathit{tz}$ \\
    
    <$RotationVar$>& $\mathit{rx}, \mathit{ry}, \mathit{rz}$ \\
    
    <$ConstVector$> & $v$ \\
    
    <$ConstString$>& $s$
\end{tabular}    
      }
    \end{mdframed}
    \vspace{-0.05in}
    \caption{Language}
    \label{fig:language}
    \vspace{-0.15in}
\end{figure}

\noindent
{\bf Language.}
Figure~\ref{fig:language} presents the language.
We call all ROS state variables $\mathit{StateVar}$,  such as {\tt tf::Point} and {\tt tf::Pose}. State variables are frame agnostic. When they are explicitly associated with some frame, they become 
$\mathit{StampedStateVar}$\footnote{ROS uses the term ``{\em stamped}'' to denote that a variable is contextualized with a frame. We hence use a similar term here.}, corresponding to {\tt tf::} {\tt Stamped$\langle$Point$\rangle$} and {\tt tf::Stamped$\langle$Pose$\rangle$}, etc. in ROS. We use $x,y,z$ for $\mathit{StateVar}$ and $\textit{fx}, \textit{fy}, \textit{fz}$ for $\mathit{StampedStateVar}$. We also use variables $\mathit{tx}, \mathit{ty}, \mathit{tz}$ and $\mathit{rx}, \mathit{ry}, \mathit{rz}$ to denote transform and rotation variables.
Note that these variables are not typed. Their types will be resolved by our type rules. We distinguish different kinds of variables just for better readability.

Statement ``$x:=v$'' denotes a state variable assignment using a constant vector. A scalar variable can be considered as a vector variable with one dimension. 
Function {\bf stamped}() explicitly associates $x$ with a frame denoted by a string $s$, corresponding to {\tt tf::Stamped$\langle$T$\rangle$::Stamped()} in ROS. 
ROS allows stamping a state variable with an empty id and later explicitly setting the id field of the stamped variable. A common error is that the developer stamps an empty frame and later forgets to set the frame.
Statement ``$\mathit{fx.id}:=s$'' models the set frame operation.
Function {\bf get\_data()} retrieves the plain state from a stamped state variable.
Function {\bf external()} models the situations where stamped state variables are published by libraries without source code and hence beyond our analysis. Function {\bf transform\_to()} transforms $\mathit{fy}$ to a target frame denoted by $s$, yielding a new stamped variable $\mathit{fx}$. It corresponds to ROS functions such as {\tt tf::Transformer\allowbreak::transformPoint()}.
Function {\bf new\_transform()} creates a one-step transform object for translating states in the child frame to their correspondences in the parent frame. 
It specifies the parent and child frame ids ( by $s_{pnt}$ and $s_{chd}$, respectively), the displacements (by state variables $x, y, z$), and the rotation (by $rx$). The displacements and rotation specify the linear and angular positions of the child frame with respect to the parent frame. 
This primitive corresponds to creating a ROS
{\tt tf::StampedTransform} object.
One-step transforms can be published 
by function {\bf sendTranform()} such that they become edges of the TF tree (Section~\ref{subsec:ros-conv}).
Function {\bf lookupTransform()} finds a transform from frame $s_{pnt}$ to frame $s_{chd}$. These two frames may not correspond to any published one-step geometric transforms. Recall that ROS automatically constructs a transform by finding a path from $s_{pnt}$ to the lowest common ancestor (in the TF tree) and then to $s_{chd}$. It corresponds to {\tt tf::Transfomer::lookupTransform()}.
Function {\bf apply\_transform()} applies a transform $\mathit{tx}$ to a stamped variable $\mathit{fy}$. As such, $\mathit{fy}$ must have the child frame type of $\mathit{tx}$ and $\mathit{fx}$ have the parent frame type of $\mathit{tx}$.
Statement ``$\mathit{rx}:=v_{3\times 3}$'' specifies a constant rotation by a 3 times 3 matrix. One can also specify a dynamic rotation from a (matrix/vector) variable. Two rotations can be aggregated to one by matrix multiplication $\otimes$. A stamped
variable can be published to a topic. Any (remote) subscribes of the topic will receive the value. As such, the variable's frame must be properly set. Otherwise, remote parties cannot make sense of it. 

While \sysname{} supports conditionals, loops, functions, and other frame related ROS APIs. They are elided from our language.

\subsubsection{\bf Type Rules}
Table~\ref{tab:rules} presents the type rules.
These rules infer types for state variables
and frame related variables and check for any inconsistencies, that is, variables cannot be properly typed, e.g., a variable having more than one types following the rules.
We use meta-variables $\ftype$, $\ttype$, and $\rtype$ to range over the infinite sets of {\it Frame}, {\it Transform}, and {\it Rotation} types. In our rules, a variable $x:\ \ftype$  means that $x$ has $\ftype$ type. A statement $x:=...:\ \ftype$ means that the statement has $\ftype$ type, which is equivalent to the left-hand-side variable $x$ having $\ftype$ type as well.
In general, a rule is read as follows: if the premises are satisfied (e.g., type checked), the conclusions are yielded. The rules are driven by
the syntax of the language.

\noindent
\underline{\em Data Assignment Rules.}
Rule $R_1$ specifies that for a copy statement, if the right-hand-side is typed to $\ftype$, the left-hand-side is typed to $\ftype$ as well.
Rule $R_2$ requires that in a binary operation of state variables, the two operands must have
the same type $\ftype$. Then we conclude the 
result of the statement has the same type.
Rule $R_3$ specifies that if $x$ is typed to $\ftype$, which has the $s$ as the frame id, we can conclude $\mathit{fx}$ has the same type $\ftype$. Note that type inference is flow-insensitive, if $x$ was free (i.e., untyped), the rule types it to $\ftype$.
Rule $R_4$ types an explicit set frame statement to the type denoted by the id string $s$.
Rule $R_5$ specifies that if a state variable 
$x$ is acquired from a stamped variable $\mathit{fx}$, $x$ inherits the type of $\mathit{fx}$. Rule $R_6$ specifies two stamped variables in a copy statement must have the same type.

\begin{figure}
\footnotesize
    \centering
    \begin{tabular}{|p{0.8in}p{2.2in}|}\hline
    $\mathit{reachable}(s_1, s_2)$: & $s_1$ and $s_2$ share some common ancestor in the TF tree and hence reachable from each other \vspace{0.05in}\\ 

    $\mathit{disp}(x)$  & = $\left\{\begin{array}{ll}
            r  & x\ \text{holds a constant}\ r\in R\ \text{after } \\ &\text{constant propagation} \\
            \top &\text{otherwise}
      \end{array}
      \right.$ \vspace{0.05in} \\
    \multicolumn{2}{|l|}{$\mathit{rotate}(<d_x, d_y, d_z>$ $,<ds_x, ds_y, ds_z>)$ = $<n_x, n_y, n_z>$, with}  \\
     \multicolumn{2}{|c|}{
    $n_x=\left\{\begin{array}{ll}
            d_x  & ds_x=``x"  \\
            opposite(d_x) & ds_x=``-x"\\
            d_y  & ds_y=``x"  \\
            opposite(d_y) & ds_y=``-x"\\
            d_z  & ds_z=``x"  \\
            opposite(d_z) & ds_z=``-x"\\
            \top & \mathit{otherwise} \\
            \end{array}
      \right.$  \ $n_y=...$,\ \ \  $n_z=...$ 
      }   \\ 
    \multicolumn{2}{|p{3.2in}|}{\vspace{0.05in}$\mathit{tree\_form\_with\_edge}(s_1, s_2)$:$\ \ $  adding an new edge denoted by} \\ 
    & $s_1\rightarrow s_2$ does not break the  TF tree form \vspace{0.05in}\\ 
    \multicolumn{2}{|p{3.2in}|}{$\mathit{orthogonalize}(v_{3\times3})$ = $\left\{\begin{array}{ll}\top &  ds_x=\top\vee ds_y=\top\vee \\
    \ & ds_z=\top \\
    \langle ds_x, ds_y, ds_z\rangle& \mathit{otherwise}\end{array}\right.$} \\
    \multicolumn{2}{|c|}{ 
       $ds_x=\left\{\begin{array}{ll}
            ``x"  & v_1=[1,0,0]  \\
            ``-x" & v_1=[-1,0,0]\\
            ``y"  & v_1=[0,1,0]  \\
            ``-y"  & v_1=[0,-1,0]\\
            ``z"  & v_1=[0,0,1]  \\
            ``-z"  & v_1=[0,0,-1]\\
            \top & \mathit{otherwise} \\
            \end{array}
      \right.$  $\ \ \ ds_y=...\ \ \ $ $ds_z=...$ 
    } \\
    \multicolumn{2}{|p{3.2in}|}{\vspace{0.05in}$\mathit{compose}(\langle ds'_x, ds'_y, ds'_z\rangle,$ $\ \langle ds''_x, ds''_y, ds''_z\rangle)$ = $\langle ds_x, ds_y, ds_z \rangle$}  \\
     \multicolumn{2}{|c|}{
    $ds_x=\left\{\begin{array}{ll}
            ds''_x  & ds'_x=``x"  \\
            -ds''_x & ds'_x=``-x"\\
            ds''_y  & ds'_x=``y"  \\
            -ds''_y & ds'_x=``-y"\\
            ds''_z  & ds'_x=``z"  \\
            -ds''_z & ds'_x=``-z"\\
            \top & \mathit{otherwise} \\
            \end{array}
      \right.$  \ $n_y=...$,\ \ \  $n_z=...$ 
      }   \\ 
    
    \hline
    \end{tabular}
    \vspace{-0.05in}
    \caption{Auxiliary functions used in type rules}
    \label{fig:definitions}
    \vspace{-0.20in}
\end{figure}

\begin{table}
  \caption{Type Rules}
  \vspace{-0.05in}
  \label{tab:rules}
  {\footnotesize
  \renewcommand{\arraystretch}{0.5}
  \begin{tabular}{L | p{0.55\columnwidth} L | L }
    \toprule
    R_1 
    & $\dfrac{y: \ftype}{x:=y: \ftype}$ 
    & R_2 
    & \dfrac{y: \ftype\ \ \ \ z: \ftype}{x:=y\ \textbf{or}\  z: \ftype} \tn \tn
    R_3
    & $\dfrac{\ftype.id=s\ \ \ \ x: \ftype}{\mathit{fx}:=\textbf{stamped}(s,x): \ftype}$ 
    & R_4
     & \dfrac{\ftype.id=s\ \ \ }{\mathit{fx.id}:=s: \ftype} \tn \tn
    R_5
    & $\dfrac{\mathit{fx}: \ftype}{x:=\textbf{get\_data}(\mathit{fx}): \ftype}$ 
    & R_6
    & \dfrac{\mathit{fy}: \ftype}{\mathit{fx}:=\mathit{fy}: \ftype} \tn \tn
    R_7
    & \multicolumn{3}{L}{
    \dfrac{\ftype.id=s\ \ \ \ \mathit{reachable}(\ftype.id, \ftype_1.id) \ \ \ \ \mathit{fy}: \ftype_1}{\mathit{fx}:=\textbf{transform\_to}(s,\mathit{fy}): \ftype}
    } \tn \tn
    R_8
    & \multicolumn{3}{L}{
    \dfrac{\splitdfrac{
      \splitdfrac{\mathit{rx}:\rtype\ \  \ftype_c=\langle s_{chd},s_{pnt},\mathit{disp}(x),\mathit{disp}(y),\mathit{disp}(z),d_x, d_y, d_z\rangle}
      {
        s_{chd}!=s_{pnt}\ \ \ \
        \ftype_p.id=s_{pnt}
      }}
    {
    \langle d_x, d_y,d_z\rangle=\mathit{rotate}(\langle\ftype_p.d_x, \ftype_p.d_y, \ftype_p.d_z\rangle, \rtype)
    }
    }
    {\splitdfrac{\mathit{tx}:=\textbf{new\_transform}(s_{chd}, s_{pnt}, x, y, z, \mathit{rx}):}{< s_{chd},s_{pnt},\mathit{disp}(x), \mathit{disp}(y), \mathit{disp}(z), \rtype>}} 
    } \tn \tn
    R_9
    & \multicolumn{3}{L}{
    \dfrac{\mathit{tx}:\ttype\ \ \ \ \mathit{tree\_form\_with\_edge}(\ttype.pid, \ttype.cid)}{\textbf{sendTransform}(\mathit{tx}): \ttype} 
    } \tn \tn
    R_{10}
    & \multicolumn{3}{L}{
    \dfrac{\mathit{reachable}(s_{pnt}, s_{chd})\ \ \ \ \ttype.pid=s_{pnt} \ \ \ \ \ttype.cid=s_{chd}} 
    {\mathit{tx}:=\textbf{lookupTransform}(s_{pnt}, s_{chd}): \ttype} 
    } \tn \tn
    R_{11}
    & \multicolumn{3}{L}{
    \dfrac{\mathit{tx}:\ttype\ \ \ \ \mathit{fy}:\ftype\ \ \ \ \ttype.pid=\ftype'.id \ \ \ \ \ttype.cid=\ftype.id}{
    \mathit{fx}:=\textbf{apply\_transform}(\mathit{tx},\mathit{fy}): \ftype'} 
    } \tn \tn
    R_{12} 
    & \multicolumn{3}{L}{
    \dfrac{\rtype=\mathit{orthogonalize}(v_{3\times 3})}{\mathit{rx}:=v_{3\times 3}: \rtype} 
    } \tn\tn
    R_{13} 
    & \multicolumn{3}{L}{
    \dfrac{}{\mathit{rx}:=x: \top} 
    } \tn\tn
    
    R_{14} 
    & \multicolumn{3}{L}{
    \dfrac{\splitdfrac{\rtype'=\langle ds'_x, ds'_y, ds'_z\rangle \ \ \ \ \rtype''=\langle ds''_x, ds''_y, ds''_z\rangle}{ \mathit{ry}: \rtype' \ \ \ \ \mathit{rz}: \rtype'' \ \ \ \ \mathit{\rtype}=\mathit{compose}(\rtype',\rtype'')}}{\mathit{rx}:=\mathit{ry}\otimes \mathit{rz}: \rtype} 
    } \tn\tn
    R_{15} 
    & \multicolumn{3}{L}{
    \dfrac{\mathit{ry}: \langle ds_x, ds_y, ds_z\rangle \ \ \ \ \mathit{rz}:\top}{\mathit{rx}:=\mathit{ry}\otimes \mathit{rz}: \langle ds_x, ds_y, ds_z\rangle} 
    } \tn\tn
    R_{16} 
    & \multicolumn{3}{L}{
    \dfrac{\mathit{fx}: \ftype}{\textbf{publish}(s_{topic}, fx):\ftype} 
    } \tn\tn

    \bottomrule
  \end{tabular}}
  \vspace{-0.6cm}
\end{table}

\noindent
\underline{\em Transformation-related Rules.}
Rule $R_7$ types a transformation statement. 
Specifically, the meaning of function $\mathit{reachable}()$ is defined in Figure~\ref{fig:definitions}. It asserts that
there is a direct/indirect transform from 
$\mathit{fy}$'s frame $\ftype_1$ to the frame $\ftype$ denoted by $s$. If so, we conclude 
that $\mathit{fx}$ has $\ftype$ type.
Rule $R_8$ types a statement that creates a one-step transform from $<s_{chd}, s_{pnt}, x, y, z, \mathit{rx}>$.
It entails the introduction of an implicit frame type $\ftype_c$ whose id is $s_{chd}$, parent id is $s_{pnt}$, displacements are those abstracted from $x, y, z$ using the $\mathit{disp}()$ function defined in Figure~\ref{fig:definitions}, and axes orientations are resulted from applying the 
(abstract) rotation specified by the rotation type $\rtype$ of $\mathit{rx}$ to the orientations of the parent frame. Function 
$\mathit{rotate}$ is defined in Figure~\ref{fig:definitions}. It takes a 3-dimension source orientation (with values {\it left}, {\it right}, {\it forward}, etc.) and a 3-dimension rotation (with values ``$x$'', ``-$y$'', etc.), and produces a resulted 3-dimension target orientation. Specifically, the 
result $x$ orientation, denoted as $n_x$ is determined by searching for which rotation dimension has value of ``$x$'' or ``-$x$''. If the $y$ rotation has value ``$x$'', denoted as $\mathit{ds}_y=``x"$, $n_x$ has the same orientation as the original $y$ orientation (i.e., $d_y$). If the value is ``-$x$'', the orientation is flipped. 
An example of such rotation can be found in the discussion of rotation type at the end of Section~\ref{subsec:types}.

Rule $R_9$ types a statement that publishes a transform
$\mathit{tx}$. It requires the edge from the parent frame to the child frame does not break the TF tree, e.g., by introducing cycles or having multiple parents for a child.
The $\mathit{tree\_form\_with\_edge}()$ function not only checks the condition but also adds the edge to the TF tree if the condition is satisfied.
Rule $R_{10}$ types a statement that looks up a transform 
from an arbitrary parent frame to an arbitrary child frame in the TF tree. The transform
may not be any of the created one-step transforms, but rather automatically composed by ROS by traversing the tree.
It requires reachability between the two frames (in the tree), which implicitly demands the existence of the two frames. Rule $R_{11}$ specifies that when typing a statement applying a transform $\mathit{tx}$ to a stamped variable $\mathit{fy}$, the transform's child frame must equal to 
$\mathit{fy}$'s frame, and its parent frame is the resulting frame. 

$R_{16}$ types a statement that publishes a stamped variable. It requires that the variable is properly typed.

\noindent
\underline{\em Rotation-related Rules.}
Rule $R_{12}$ types the creation of a rotation from a $3\times 3$ constant vector. Function $\mathit{orthogonalize}()$ is used to abstract the vector to the corresponding $\mathit{DirSwitch}$ abstract values. Its definition can be found in Figure~\ref{fig:definitions}. 
Specifically, an orthogonal $3\times 3$ vector is composed of three orthogonal unit vectors, i.e., each row and each column has only one non-zero element of value 1 or -1. Now, if we consider that rows 1,2,3 represent initial $x,y,z$ orientation and columns 1,2,3 represent then changed $x,y,z$ orientation, then the column position of the non-zero element in a row indicates the axis of the changed orientation that is the same as the axis of the initial orientation represented by the row. For example, value -1 in row 2 and column 1 denotes that the direction of $y$ axis of initial orientation becomes the direction of $-x$ axis of the changed orientation. When the matrix does not denote orthogonal rotation, the resulted type is $\top$.

Rule $R_{13}$ types a rotation creation
from a variable (i.e., non-constant). 
It usually corresponds to angular transformation that is needed for state translation across frames, but not for orientation changes. For example, if a camera can spin regarding the body, its readings (regarding the camera frame) have to go through a dynamic angular transformation in order to be interpreted regarding the body frame.  
Note that in this case, the angular transformation is independent of frame orientations. Both the 
camera and the body frames have orthogonal orientations. 
Rule $R_{14}$ types a rotation composition statement. 
It leverages function $\mathit{compose}()$ to derive a new rotation type.
In the function definition in Figure~\ref{fig:definitions}, $\langle ds'_x,ds'_y,ds'_z\rangle$ and $\langle ds''_x,ds''_y,ds''_z\rangle$, both denote change in the orientation, where the former change is followed by the latter. For example, {\tt Rz} and {\tt Rx} vectors in our motivating example (Figure~\ref{fig:motiv}) represent $\langle -y,x,z\rangle$ and $\langle x,-z,y\rangle$ changes respectively as per rule $R_{12}$. Part A in the below figure shows these orientation changes in a sequence; whereas part B shows the orientation change that is computed by aggregating these two changes. Note that both parts result into the same orientation.

\begin{center}
    \includegraphics[width=0.7\columnwidth]{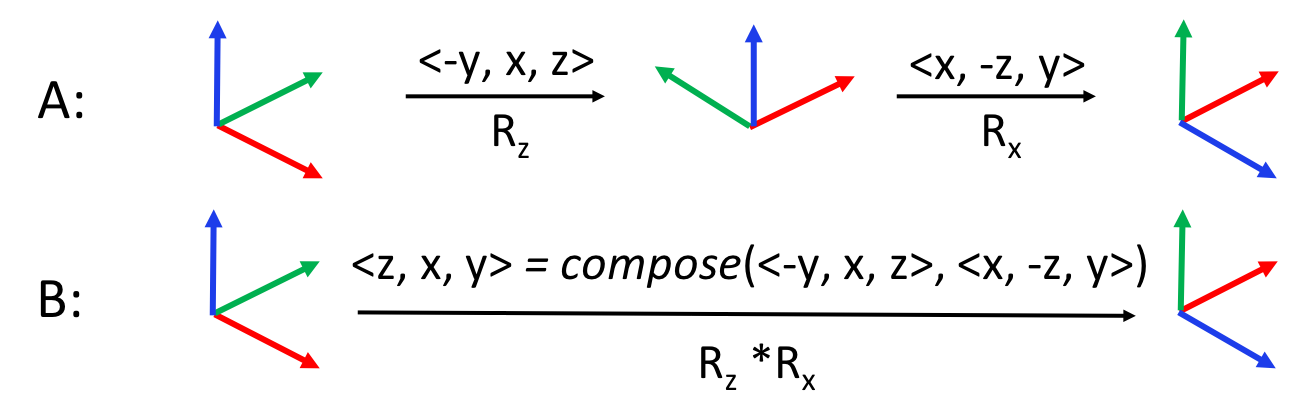}
\end{center}
\vspace{-0.10in}

$R_{15}$ types a rotation composition in which one of the rotations is not static or not orthogonal. Note that an arbitrary angular rotation can be integrated with an orthogonal axes orientation rotation.
Since we only model and check axes orientation, the resulted type is the 
one that represents orthogonal orientation changes. 

\begin{table}
  \caption{Type checking of example in Figure~\ref{fig:motiv}}
  \vspace{-0.05in}
  \label{tab:rules-on-example}
  {\footnotesize
  \begin{tabular}{L | L | L | L }
    \toprule
    \text{St\#} & \text{Statement} & \text{Type} & \text{Rule}
    \tn \midrule
    6 
    & \mathit{tf\_cam\_rotation}:=\mathit{tmp}
    & \top
    & R_{13}
    \tn
    10 
    & \mathit{Rx}:= [[1,0,0],[0,0,-1],[0,1,0]]
    & \langle x,-z,y\rangle
    & R_{12}
    \tn
    13
    & \mathit{Rz}:= [[0,-1,0],[1,0,0],[0,0,1]]
    & \langle -y,x,z\rangle
    & R_{12}
    \tn
    20
    & \mathit{tf\_cam\_rotation}:=\mathit{Rx} \otimes \mathit{tf\_cam\_rotation}
    & \langle x,-z,y\rangle
    & R_{15}
    \tn
    21
    & \mathit{tf\_cam\_rotation}:=\mathit{Rz} \otimes \mathit{tf\_cam\_rotation}
    & \langle z,x,y\rangle
    & R_{14}
    \tn
     23
     & \mathit{tf\_cam\_rotation}:=\mathit{invYZ} \otimes ...
     & \langle z,-x,-y\rangle
    & R_{14}
     \tn
    25
    & \mathit{rtmp}:= \mathit{tf\_cam\_rotation}
    & \langle z,-x,-y\rangle
    &  
    \tn
    30
    & 
    \mathit{ttmp}:= \textbf{new\_transform}(s_{\mathit{cam}}, s_{\mathit{map}},...,\mathit{rtmp})
    & 
    \rr \makecell{\text{error because }\\
    \textbf{rotate}(\langle f, l, u\rangle, \\ \langle z, -x, -y\rangle)=  \\ \langle r, d, f\rangle}
    & R_8
    \tn
    \bottomrule
  \end{tabular}}
  \vspace{-0.5cm}
\end{table}

\noindent\underline{Example.}
Table~\ref{tab:rules-on-example} presents part of the example code in Figure~\ref{fig:motiv} written in our language and the types computed by the
type rules. 
The first column shows the original line numbers of the statements. 
St\# 6 is an assignment of a rotation variable from a state variable. The state variable \textit{tmp} gets a value from the external value state variable \textit{rotation}. Therefore, as per rule $R_{13}$,  \textit{tf\_cam\_rotation} is typed to $\top$. St\# 10 and 13 are the assignments of rotation variables from constant vectors. Rule $R_{12}$ allows typing \textit{Rx} and \textit{Rz} to $\langle x,-z,y\rangle$ and $\langle -y,x,z\rangle$, respectively.
Further, St\# 20 and 21 denote the composition of rotations: the former is typed with the type of \textit{Rx} by $R_{15}$ as the type of \textit{tf\_cam\_rotation} is $\top$, whereas the latter is typed with $\langle z,x,y\rangle$, which is computed by composing types of \textit{Rz} and \textit{tf\_cam\_rotation} using rule $R_{14}$\footnote{\sysname{} uses SSA to handle variable re-definitions.}.
St\# 23 and 25 are similarly typed.
Finally, St\# 30 constructs a new transform object.
However, it cannot be properly typed by rule $R_8$ as applying
rotation $\langle z,-x,-y\rangle$
to the parent map frame with the conventional orientation  $\langle\textit{forward},\textit{left},\textit{up}\rangle$, i.e, FLU, yields 
the orientation 
$\langle\textit{right},\textit{down},\textit{forward}\rangle$, contradicting with the ROS orientation convention FLU for the camera frame. On the other hand, \sysname{} correctly type-checks the fixed version.

\noindent
{\bf Soundness and Completeness.}
The essence of \sysname{} is to leverage  flow-insensitive static analysis (like type checking) to find frame related bugs. It is automated and does not require any user type annotations
for subject ROS projects. As such, the type
system is neither sound nor complete. This is because some type rules are uncertain. For example in rule $R_{12}$, a matrix representing orthogonal 
angular transformation may not be intended for frame axes orientation changes, but rather because the parent and child body parts are installed in a way that they form some orthogonal angle(s), while their frame axes orientations are identical (both FLU). 
However, since there is no way for \sysname{} to know the real meanings of such rotation without user input, it considers this an axes orientation rotation, which is the most common case.
And stamped variables from third party libraries are widely used and \sysname{} may not be able to type them. Therefore, \sysname{} has both false positives and false negatives.
On the other hand, we foresee future robotic system development frameworks may support full-fledged frame types which the developer can use to declare their variables. With such explicit type declarations, a sound and complete type system can be developed.  


\subsubsection{\bf Frame Related Problems.} 
\label{sec:typeerrors}
\sysname{} reports 7 kinds of type errors, 2 kinds of type warnings and 3 kinds of violations of implicit conventions. The first two are also called inconsistencies. 

\noindent
{\bf Type Errors.}
This first kind is 
{\em TF tree related errors.} These errors include {\em a frame having multiple parent frames}, {\em frames forming a cycle}, and {\em frames not following conventional orders} such as a map frame must be the ancestor of a body frame (Section~\ref{subsec:conventions}).

The second kind of type errors is {\em incorrect frames or transforms.}
ROS allows developers to stamp a state variable $x$ with an empty id.
If these variables are published, meaning that other components from different parties may receive such data, they should be stamped with a proper frame (rule $R_{16}$). Moreover, if they are used in transforms, runtime errors will be triggered as well.
In these cases, \sysname{} reports a {\em missing frame error}.  
Although not explicitly modeled by our rules, a few ROS data types require explicitly specifying child frame ids.
If child frames are not set, \sysname{} reports {\em missing child frame errors}.

A transform is required to have different parent and child frames (rule $R_8$). Otherwise, it is meaningless. \sysname{} reports a
{\em redundant transform}.
Rule $R_8$ requires that the displacements and rotation in a newly created transform should yield (relative) positions and axes orientations consistent with the parent frame and ROS conventions (e.g., body frames should have {\em forward-left-up} (FLU) orientations). Otherwise, it reports
 {\em incorrect transforms}.

\noindent {\bf Type Warnings.}
A sensor transform that defines a sensor frame with respect to some other robotic body part like base\_link, wrist, etc. or vice versa should have at least some non-zero displacement(s)/rotation(s) as a sensor's center unlikely aligns with a body part's center. \sysname{} reports a warning when a sensor transform has zero displacements and rotations. 
For a transform from parent to child, ROS naming conventions follow templates `<parent>\_to\_<child>' and `<child>\_in\_<parent>', which give the clear meanings of the transform.
Developers however often inverse the parent and the child in transform variable names. \sysname{} hence reports {\em reversed name warnings}. These two are warnings as they may not affect the functionalities but rather degrade maintainability. Such checkings are not explicitly modeled in type rules although \sysname{} supports them.

\noindent {\bf Implicit Convention Violations.}
As mentioned in Section ~\ref{subsec:conventions}, implicit conventions are mined from launch files. As such, they may not be required but rather commonly seen. \sysname{} reports three kinds of warnings. The first is {\em null displacement expected} warning, meaning that certain transforms or frames are supposed to have null displacements. The second is {\em null rotation expected} warning, meaning that certain transforms 
are supposed to have null rotations. The third is that {\em co-occurrence violation} warnings when frames that are commonly seen together (in other projects) are not present in a project. Such checkings are performed as an additional step after frame and transform types are inferred. 
The reason is that it is hard to implement implicit conventions as type rules as the mined conventions may change depending on the repositories used and the z-score threshold setting.

\begin{table*}
\captionsetup{belowskip=0pt,aboveskip=5pt}
\caption{Inconsistencies by Category (with the first seven type errors, the last two type warnings)}
\label{tab:type}
  {\footnotesize
  \begin{tabular}{ p{0.3\textwidth} | p{0.05\textwidth} | p{0.05\textwidth} p{0.05\textwidth}| p{0.05\textwidth} p{0.05\textwidth}| p{0.05\textwidth} p{0.05\textwidth} p{0.05\textwidth} } 
    \toprule
    \multicolumn{1}{p{0.3\textwidth}}{Category} & 
    \multicolumn{5}{|p{0.25\textwidth}}{\rr Number of Inconsistencies} & 
    \multicolumn{3}{|p{0.15\textwidth}}{\rr Number of Repositories} \\
    \cmidrule(l){2-6} \cmidrule(l){7-9}
    & Total (\#)
    & \multicolumn{2}{|p{0.1\textwidth}}{\rr TP (\#)} & \multicolumn{2}{|p{0.1\textwidth}|}{\rr FP (\#)}
    & {\rr Total (\#)} & {\rr TP (\#)} & {\rr FP (\#)} \\
    \cmidrule(l){3-6}
    & & C/C++ & Launch & C/C++ & Launch & & & \\
    \midrule
    c\_MULTIPLE\_PARENTS\_IN\_FRAME\_TREE & 4 & 3 & 1 & 0 & 0 & 4 & 4 & 0 \tn
    c\_CYCLE\_IN\_FRAME\_TREE & 1 & 1 & 0 & 0 & 0 & 1 & 1 & 0 \tn
    td\_INCORRECT\_FRAME\_ORDER\_IN\_TREE & 7 & 3 & 4
    & 0 & 0 & 7 & 7 & 0 \tn
    c\_INCORRECT\_TRANSFORM & 17 & 4 & 4 & 0 & 9 & 9 & 3 & 6 \tn
    td\_REDUNDANT\_TRANSFORM & 2 & 2 & - & 0 & - & 2 & 2 & 0 \tn
    td\_MISSING\_FRAME & 108 & 92 & - & 16 & - & 56 & 49 & 13 \tn
    td\_MISSING\_CHILD\_FRAME & 12 & 8 & - & 4 & - & 7 & 4 & 3 \tn
    td\_REVERSED\_NAME (type warning) & 33 & 1 & 28 & 1 & 3 & 19 & 17 & 4 \tn
    c\_SENSOR\_NULL (type warning) & 6 & 0 & 3 & 0 & 3 & 5 & 2 & 3 \tn
    \midrule
    Total & 190 & 114 & 40 & 21 & 15 & 110 & 89 & 29 \tn
    & & [81.05\%] & & [18.95\%] & &  &  & \tn
    \bottomrule
  \end{tabular}}
\end{table*}

\begin{table}
\captionsetup{belowskip=2pt,aboveskip=5pt}
\caption{Summary of True Positive Inconsistencies
} 
\label{tab:detail}
  {\footnotesize
  \renewcommand{\arraystretch}{0.98}
  \begin{tabular}
    { p{0.277\columnwidth} | p{0.052\columnwidth} | p{0.528\columnwidth} }
    \toprule
     Report \& Response & Count & Inconsistency Categories \\
    \midrule
    \multirow[t]{3}{=}{\rr found by searching commit-history or issue reports of repositories.}
    & 3 & td\_MISSING\_FRAME \tn
    & 1 & c\_INCORRECT\_TRANSFORM \tn 
    & & \tn 
    \midrule
    \multirow[t]{4}{=}{\rr fixed by developers before we reported them.}
    & 1 & td\_MISSING\_FRAME \tn
    & 1 & td\_REDUNDANT\_TRANSFORM \tn
    & 1 & c\_CYCLE\_IN\_FRAME\_TREE \tn  
    & 1 & c\_MULTIPLE\_PARENTS\_IN\_FRAME\_TREE \tn \midrule
    \multirow[t]{4}{=}{\rr reported, and either fixed by developers or ready to accept a pull-request.}
    & 2 & td\_MISSING\_FRAME \tn
    & 1 & td\_MISSING\_CHILD\_FRAME \tn
    & 1 & td\_REDUNDANT\_TRANSFORM \tn 
    & 1 & td\_INCORRECT\_FRAME\_ORDER\_IN\_TREE \tn
    \midrule
    \multirow[t]{7}{=}{\rr reported, and acknowledged by developers (in some cases, confirmed as a useful fix to others for reusing the package node).}
    & 1 & td\_MISSING\_FRAME \tn
    & 5 & td\_REVERSED\_NAME \tn
    & 2 & c\_SENSOR\_NULL \tn
    & 2 & td\_INCORRECT\_FRAME\_ORDER\_IN\_TREE \tn
    & & \tn
    & & \tn
    & & \tn
    \midrule
    \rr reported, but refused by developers on the grounds that reuse of the package node is not easy, or the package is no longer maintained.
    & 3 & td\_MISSING\_FRAME \tn 
    \midrule
    \multirow[t]{6}{=}{\rr reported, but no response from developers yet.}
    & 12 & td\_MISSING\_FRAME \tn 
    & 1 & td\_MISSING\_CHILD\_FRAME \tn
    & 4 & c\_INCORRECT\_TRANSFORM \tn 
    & 14 & td\_REVERSED\_NAME \tn
    & 1 & td\_INCORRECT\_FRAME\_ORDER\_IN\_TREE \tn
    & 2 & c\_MULTIPLE\_PARENTS\_IN\_FRAME\_TREE \tn \midrule
    \multirow[t]{7}{=}{\rr found in a package node that is reused by other developer. The issue is either fixed or acknowledged by other users in the original repository of the node.}
    & 2 & td\_MISSING\_FRAME \tn 
    & 4 & td\_MISSING\_CHILD\_FRAME \tn
    & & \tn
    & & \tn
    & & \tn
    & & \tn
    & & \tn
    \midrule
    \multirow[t]{3}{=}{\rr found in backup-copy files 
    in repositories.}
    & 5 & td\_MISSING\_FRAME \tn 
    & & \tn
    & & \tn
    \midrule
    \rr potential issues.
    & 8 & td\_MISSING\_FRAME \tn 
    \midrule
    \multirow[t]{7}{=}{\rr projects not active in past two years.}
    & 55 & td\_MISSING\_FRAME \tn
    & 2 & td\_MISSING\_CHILD\_FRAME \tn 
    & 3 & c\_INCORRECT\_TRANSFORM \tn
    & 10 & td\_REVERSED\_NAME \tn
    & 1 & c\_SENSOR\_NULL \tn
    & 3 & td\_INCORRECT\_FRAME\_ORDER\_IN\_TREE \tn
    & 1 & c\_MULTIPLE\_PARENTS\_IN\_FRAME\_TREE \tn
    \bottomrule
  \end{tabular}}
\end{table}

\begin{table}
\captionsetup{belowskip=2pt,aboveskip=5pt}
\caption{Implicit Convention Warnings by Category}
\vspace{-0.05in}
\label{tab:warnings}
  {\footnotesize
  \begin{tabular}
  { p{0.285\columnwidth} | p{0.04\columnwidth} p{0.03\columnwidth} | p{0.04\columnwidth} p{0.03\columnwidth} | p{0.04\columnwidth} p{0.03\columnwidth} | p{0.04\columnwidth} p{0.03\columnwidth} }
    \toprule
    \multicolumn{1}{p{0.285\columnwidth}}{Category} & 
    \multicolumn{2}{|p{0.07\columnwidth}}{\rr z=1} & 
    \multicolumn{2}{|p{0.07\columnwidth}}{\rr z=2} &
    \multicolumn{2}{|p{0.07\columnwidth}}{\rr z=5} &
    \multicolumn{2}{|p{0.07\columnwidth}}{\rr z=10} \\
    \cmidrule(l){2-9}
    & W\# & R\# & W\# & R\# & W\# & R\# & W\# & R\#\\
    \midrule
    w\_sig\_null\_rot\_exp & 18 & 9 & 16 & 7 & 14 & 5 & 9 & 2 \tn
    w\_name\_null\_rot\_exp & 20 & 10 & 20 & 10 & 2 & 2 & 0 & 0 \tn
    w\_name\_co-occurrence & 7 & 3 & 0 & 0 & 0 & 0 & 0 & 0 \tn
    \bottomrule
  \end{tabular}}
  \vspace{-0.4cm}
\end{table}

\section{Evaluation}
\label{sec:eval}

We address
the following questions:
{\bf RQ1.} How effective and efficient is \sysname{} in disclosing frame related inconsistencies? 
{\bf RQ2.} What is the response from developers on the inconsistencies reported by \sysname{}?
 {\bf RQ3.} What kinds of implicit conventions are violated and what is the impact of z-score on results?

\subsection{Experiment Setup}

\textbf{Implementation}. We have implemented a prototype of \sysname{} in around 4800 lines of Python code. It utilizes a third-party component, \emph{cppcheck}~\cite{cppcheck}, to obtain an XML dump for a C/C++ file providing an intermediate representation. It contains program tokens, symbol tables for scopes, functions and variables, and an AST for each statement. We have also implemented constant propagation to resolve variables to static values when possible. The miner is implemented using MySQL database and SQL script.
Implementation details can be found in Appendix C
\iflong \else in~\cite{FullVersion}\fi. 
The database for the implicit convention miner is built from a list of ROS projects that contain at least one launch file with a \emph{static\_transform\_publisher}. We use only the mature projects from this list. We consider a project mature when it has more than 30 commits. 
The database contains 12.1K static transforms in 5.3K launch files from 2.2K mature projects.
The tool and data is available at 
\url{https://doi.org/10.5281/zenodo.4959920}

\noindent
\textbf{Artifacts}. We run the type checker on 180 ROS projects from Github.
They include (a) 7 projects from the ROS Kinetic 
distribution that 
contain at least one issue or a commit related to frame; (b) 23 randomly selected projects from the ROS indigo distribution; and (c) 150 randomly selected ROS projects 
that contain at least one static transform.
Among these projects, 69 are mature and 99 not (with less than 30 commits). Note that we consider evaluation on unmature projects important as well because they are likely by non-expert ROS developers that can benefit more from \sysname{}.
Table 7
in Appendix 
\iflong \else in~\cite{FullVersion} \fi 
shows the 
statistics of these projects. On average,
each project has 45 source/launch files and 52491 
LOC, with the largest one having 433 files and over 7 million LOC.  

\noindent
\textbf{Examination Process}. We examine each of the reported inconsistencies 
by reviewing the corresponding source code and mark it as \emph{True Positive} (TP) or \emph{False Positive} (FP). The manual examination poses a threat to the validity of results presented in this paper (please refer to Appendix D
\iflong \else in~\cite{FullVersion}\fi).
We mitigate it by cross-checking, being very conservative in marking TPs, and confirmation with developers.
The implicit convention violation warnings are by their nature more difficult to determine if they denote real bugs.  
We hence run the tool on the same set of projects with different z-score threshold values and study the reported warnings.
All type checking runs are on a Dell PowerEdge R420 Linux server with two Intel Xeon E5-2690 2.90GHz 8-core processors and 192 GB of RAM. 

\subsection{Results and Observations}
\noindent
{\bf RQ1. Effectiveness and Efficiency.} \sysname{} reports 190 inconsistencies with 154 true positives and 81.05\% true positive rate. These inconsistencies belong to 100  projects, over 55\% of the total we examined. This number is specially significant considering that many more frame related bugs reported in earlier versions of the systems may have been detected by \sysname{}. 
\sysname{} also reports 45, 36, 16, 9 implicit convention violation warnings when the z-score threshold is set to 1, 2, 5, 10, respectively, which corresponds to p-values smaller than 0.15 for z=1 to p-values smaller than 0.00001 for z=10.
The average analysis time is 52 seconds  
with the maximum 1190 seconds. Further details can be found in 
Table 6
in Appendix 
\iflong \else in~\cite{FullVersion}\fi. 


\noindent
{\em Inconsistencies by Category.}
Table \ref{tab:type} summarizes inconsistencies by their category, with the first seven type errors and the last two type warnings. The first column lists the categories. Each has either a `$c\_$' or `$td\_$' prefix, denoting whether it is a ``correctness'' or a ``technical debt'' issue, respectively. The ``correctness'' issues represent the types of inconsistencies that violate ROS conventions such as a tree representation of all the frames in an application, sensor frames should have non-null transforms with respect to robot body frames, and incorrect transforms. The incorrect transform issues include those that have invalid displacement/rotation (like our motivating example in Section~\ref{sec:motivation}) which can lead to problems during runtime, and those that do not follow ROS naming conventions (discussed in Section~\ref{subsec:ros-conv}) while defining a child frame in the transform which affect code readability and may cause issues during code reuse. The ``technical debt'' issues represent the types of inconsistencies that may make code maintenance and understanding difficult and may lead to potential runtime problems during code reuse/collaboration. The next five columns show the total number of inconsistencies for each category, TP cases in C/C++ code and in launch files, FP cases in C/C++ code and in launch files, respectively. Further, the last three columns show the number of projects with inconsistencies, TPs, and FPs, respectively.

Many of the reported inconsistencies are about missing frames, i.e., developers forgot
to set frame ids for published state values. 
FP cases of this type are observed due to usage of stamped data types for storing non-frame data (e.g., binary mask of image), usage of stamped transform for tf operations that do not need \emph{frame\_id} field value, and passing of stamped data type variables to functions whose signature is not available during analysis. Similar sources of false positives affect other 
categories. \sysname{} also found many reversed name warnings. For example, ROS expects transform variables to follow the naming convention of ``$\langle$parent$\rangle$\_to\_$\langle$child$\rangle$''. We find that many developers use the names reversed. This may cause maintenance and reuse issues as the convention is expected by other parties.
\sysname{} also identified 17 frame type inconsistencies similar to our motivating example in Section~\ref{sec:motivation}
and 
12 TF tree errors.
Given that ROS is built to encourage collaborative development, even the issues affecting code readability or reuse can be important.
We have three case studies in Appendix E
\iflong \else in~\cite{FullVersion}\fi. 


\noindent
{\bf RQ2. Developer Responses.}
Table~\ref{tab:detail} summarizes the distribution of the true positives across different reporting actions, developer responses and inconsistency categories.
Row 1 (below the table title) shows that 4 of the 154 true positives \sysname{}  uncovered were known issues by the developers.
Row 2 shows that another four were  fixed by the time we reached the  developers. 
Among the remaining true positives, we reported the 52 that belong to projects that showed activity in the past 2 years (rows 3-6).  We received 18 responses with 15 either acknowledging this issue or fixing it, and 3 declining our reports because either it is not seen as relevant (project is not intended for external use hence conventions do not need to be respected), or project no longer maintained.
Row 8 shows that \sysname{} found 6 bugs  in reused versions of projects that have already been fixed in the original repositories.

\noindent
{\bf RQ3. Implicit Convention Violations.}
Table \ref{tab:warnings} shows the three kinds of implicit convention violations \sysname{} has found and the impact of z-score threshold.
W\# means the number of warnings and  R\# the number of projects with warnings.
The most popular kind is {\em w\_name\_null\_rot\_exp}, meaning that a transform with a specific name is expected to have 0 rotation; {\em w\_sig\_null\_rot\_exp} is similar except that a transform is identified by parent and child frame ids instead of name.
Some of the {\em w\_sig\_null\_rot\_exp} warnings are persistent (with different z-threshold), meaning that conventions adopted by almost all projects are being violated.

\vspace{-0.2cm}
\section{Related Work}
\label{sec:relwork}
\noindent
{\bf Component-based Robot Software.} 
Component-based programming~\cite{brooks2005towards, brugali2009component} has been extensively used in robot development, aiming to facilitate reuse.
Following the principle, several robot software frameworks~\cite{magyar2015comparison}, such as Open Robot Control Software (Orocos)~\cite{bruyninckx2001open}, ROS~\cite{quigley2009ros}. Player~\cite{gerkey2003player}, Miro~\cite{utz2002miro}, have been developed and widely used. 
These frameworks commonly provide communication infrastructure across components~\cite{rostopics} and substantial library support. 
As such, respecting conventions is critical for these infrastructures. Unfortunately, they provide no compile time support to ensure frame conventions, one of the most error-prone aspects of robot programming.
\sysname{} is a  
type checking technique that systematically detects problems in using frames. 

\noindent
{\bf Type Checking in Robotic Systems.} 
Type checking~\cite{reynolds1974towards, cardelli1996type} is a well-known technique to improve software reliability. 
In robotic software, due to its domain-specific characteristics, variables often have physical-world semantics. Additional type checking can hence be performed considering the physical world.  
Unit type checking~\cite{biggs2007designing, damevski2009expressing} is the most common robotic software type system, which type variables with physical unit information. 
The seminal work~\cite{kennedy1994dimension, rittri1995dimensioninference, kennedy2009unitsofmeasure} in the type checking of units and dimensions extends a programming language.
Ayudante~\cite{haq2015ayudante} builds and compares clusters of variables based on dataflow and the meaning of variables names to detect type (e.g., unit) inconsistencies.
The recent work on Phriky  \cite{Ore:2017:LDP:3092703.3092722, Ore2017Phriky} supports dimensionality type checks in ROS without developer annotations, and Phys~\cite{kate2018phys} extends it with probabilistic type inference to detect more inconsistencies. In this work we extend the type checking to reference frames in robotic software, 
which requires a distinct type system. 

\noindent
{\bf Frame Safety.}
Lowry et al.~\cite{lowry2001certpolicies} proposed a general abstract interpretation system for certifying domain-specific properties and provide a case study to certify NAIF library programs’ frame-safety. The case study checks limited properties (e.g., frame consistencies) and the system needs annotations for program inputs. In contrast, ours is a fully automated tool checking many properties for ROS projects (e.g., TF tree shape and frame orientations), without requiring annotations. Our abstract domains are hence different from theirs and our method is type system based. 
The work by Basir et al.~\cite{basir2010hierarchicalsafety} on automated theorem proving for automatically generated code also presented a case study to verify two frame-safety requirements. 
META-AMPHION system~\cite{lowry1997metaamphion} helps domain experts build  domain-specific program synthesizers. In the context of frame-safety, we believe that this work would enable the synthesis of programs that are frame-safe given the correct domain theory. 
In our work, we focus on finding frame issues in the already developed code.
Overall we do not need annotations, have a push-button approach, and provide a broad assessment of real code.

\vspace{-0.2cm}
\section{Conclusion}
\label{sec:conclusion}
We develop a 
fully automated type checking technique for reference frames in robotic system, which is one of the most subtle and error-prone aspects in robotic software engineering. We explain the semantics of frames and frame transformations, and propose a novel way to abstract them. Type checking rules are developed based on the domain specific semantics. Applying our technique to 180 ROS projects from GitHub discloses over 190 errors and warnings, with 154 true positives. We reported 52 of them and received 18 responses by the time of submission and 15 fixed/acknowledged.

\vspace{-0.2cm}
\begin{acks}
We thank our insightful reviewers.
This research was supported, in part by NSF 1901242, 1910300, 1909414 and 1853374, ONR N0001417\-12045, N000141410468 and N000141712947. Any opinions, findings, and conclusions in this paper are those of the authors only and do not necessarily reflect the views of our sponsors.
\end{acks}

\bibliographystyle{ACM-Reference-Format}
\balance
\bibliography{main}

\iflong
\clearpage

\appendix
\section*{Appendix}
\label{sec:appendix}
\nobalance

\begin{figure}[!htbp]
    \centering
    \includegraphics[width=0.95\columnwidth]{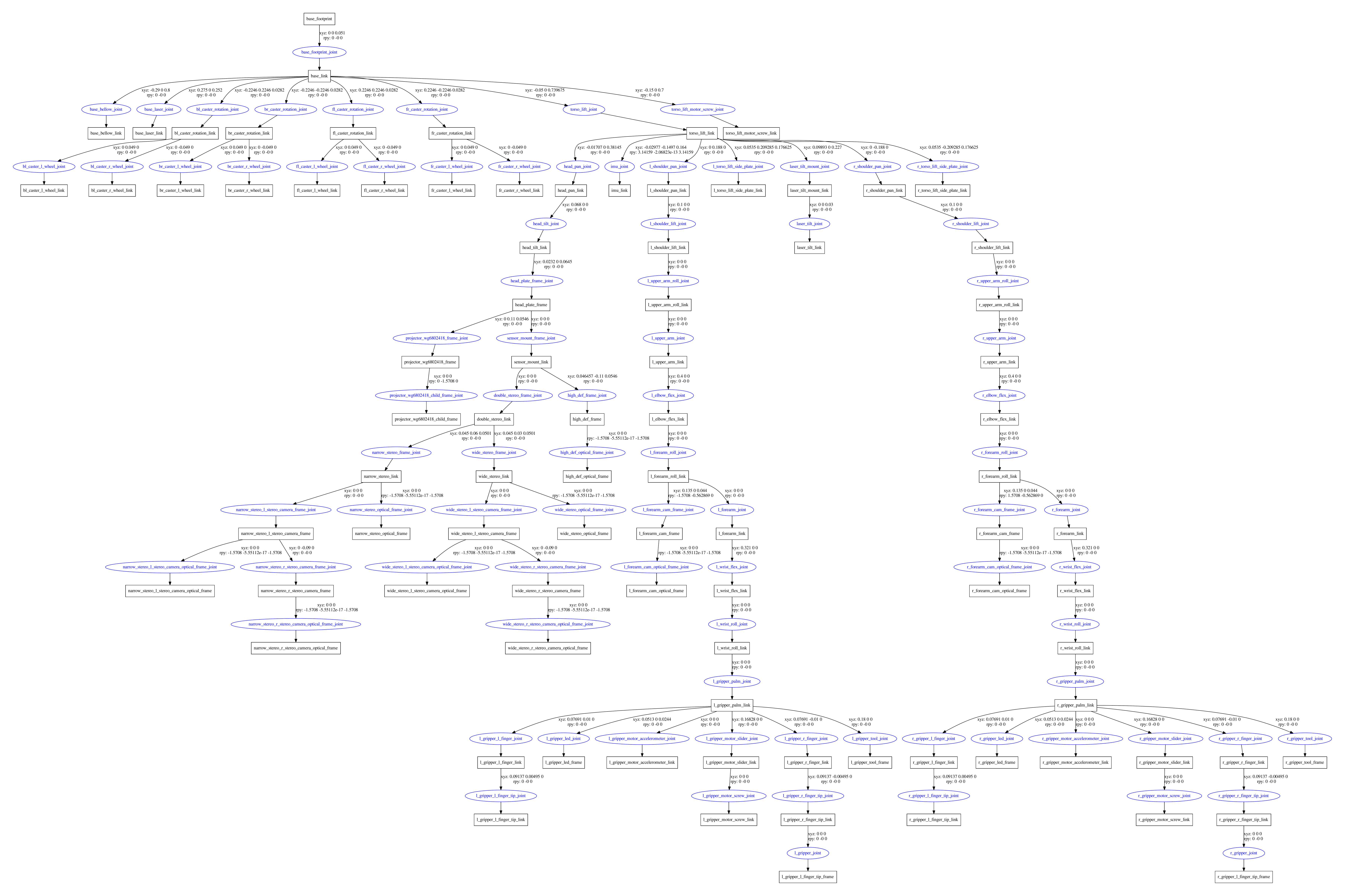}
    \caption{PR2 Robot Coordinate Frames (a square node represents a frame)}
    \label{fig:pr2_tree}
\end{figure}

\section{An Example Launch File Snippet}
\label{app:launch}
The following shows a snippet from a launch file defining a transform from a {\it base\_link} 
frame to an optical depth frame for a Robotiq gripper simulation~\cite{launch}.  
 
\begin{lstlisting}[basicstyle=\footnotesize\ttfamily, columns=fullflexible, breaklines=true]
  <node name="kinect_link_broadcaster" pkg="tf" type="static_transform_publisher" args="0.536 0 1 1.5708 3.14159 0 base_link openni_depth_optical_frame 100" />
\end{lstlisting}
The {\em name} field uniquely identifies the transform and the {\em args} field provides parameters of the transform 
with the first three values the $x$, $y$, $z$ displacements of the origin of the child frame regarding the parent frame; the next three values the yaw, pitch, roll rotation angles; followed by the parent and child frame ids.

\section{Implicit Convention Miner}
\label{app:miner}
The miner is only executed once on the collected static transform database. The produced results are then used in the analysis for each subject ROS project. It identifies the following information.

\begin{itemize}[leftmargin=*]
    \item Pairs of transforms that commonly co-exist in a project. For example, \sysname{} finds that the transform {\it base\_link}$\rightarrow$ {\it left\_wheel} co-exists with {\it base\_link}$\rightarrow$ {\it right\_wheel}. As such, violations to this implicit convention may potentially be buggy.
    
    \item  Transforms commonly having zero displacements, such as
    {\it kinect\_link}$\rightarrow${\it kinect\_depth\_frame}.
    
    \item Transforms commonly having zero rotation, 
    such as {\it world}$\rightarrow${\it map}. 
\end{itemize}

Since these patterns may happen coincidentally, we use the following z-score\cite{z-score} to quantify their certainty. 
\begin{equation*}
    z(n,e) = (e/n-p_0)/\sqrt{(p_0*(1-p_0)/n)}
\end{equation*}
For a relation (pattern) denoted as a pair of premise and conclusion, $n$ 
is the number of times the premise  is observed, $e$ is the number of times
both the premise and conclusion are observed. 
For example in computing the z-score for a transform $A$ having zero rotation, $n$ is the number of times $A$ is observed and $e$ is the number of times $A$ having zero rotation.
Coefficient $p_0$ denotes the expected likelihood that a conclusion is satisfied.
We use $p_0=0.9$ following the literature. Note that $z$ grows as $n$ increases and $n-e$ decreases.

\section{Prototype Implementation}
\label{app:implementation}
Prototype implementation of \sysname{} contains around 4800 lines of python code and 160 lines of SQL script. It can be invoked on command line with the project path input.

The implicit convention miner uses MySQL to store the static transforms (present as \emph{static\_transform\_publisher} nodes in launch files) extracted from a list of 6K projects. The list was generated using the GitHub code search API which followed: downloading of 55K launch files that contained a \emph{static\_transform\_publisher} from public GitHub repositories; downloading of an original version of each file that was found to be modified in the commit history of its GitHub repository, removal of files which could not be parsed correctly, removal of duplicate files as the launch files are often shared between projects. Each row in the database represents a static transform with the following fields: a launch file id, a project id, each of the attributes and arguments of the \emph{static\_transform\_publisher} node, a flag indicating the version of the launch file (original or later commit), a flag indicating if the project is mature (more than 30 commits). Note that we skip static transforms that employ macro expansion in the node arguments (which we leave it as a future task). In total our database contains 26K static transforms from 13K files in 6K projects, out of which 12.1K static transforms are from mature projects. The miner extracts conventions from the 12.1K static transforms. 

Further, the miner uses SQL scripts to build a table of observed relations (patterns) for each of the kinds discussed in Appendix~\ref{app:miner}, and to compute the $z$ scores of those relations. Then, given the z-score threshold $t$, it outputs the relations with $z>t$ 
as implicit conventions. 
Note that the execution of the miner is a one time step and is not run each time the tool is invoked on a ROS project to be type-checked.

When \sysname{} is invoked on a subject project, the tool first executes the file-processor component to obtain a group of C/C++ and launch files that are executed together for each of the launch files in the project. A C++ ROS project contains a set of packages, a set of nodes in each package and, optionally, a set of launch files which invoke nodes and other launch files. A node can also be invoked on its own. A node is specified using a package name and an executable name. Therefore, it processes package.xml files to obtain the package names available in the project, and CMakeLists.txt files to obtain the sets of C/C++ files belonging to executables. We use the \emph{cmakelists\_parsing} python module as a CMakeLists.txt file parser. Note that the file-processor considers only those nodes that are present in the project. A launch file may invoke other project’s nodes, e.g., ROS core nodes. These file groups are used to construct valid TF trees.

Further, recall that our type system encodes the displacement fields in the types with static numerical values if any. For that, we implement constant propagation to be performed before the type checking. The constant propagation at the time of this work is implemented at a function level, where it follows a visitor pattern to annotate numerical constants and variables with their values: the values can be either a number or a numerical vector; then to propagate the annotated values on the AST of each statement in the function.

Finally, the type checker component applies the type rules on the variables and statements of each C/C++ file present in the project. The implementation supports the transfer of types across functions: from a function's return type to the caller, and from a caller to the function parameters. At the time of this work, this transfer happens between functions at the file level. The transfer of types between functions can be extended at the file groups level, which we leave it as a future task. Note that however the TF trees constructed through the type rules are maintained at the file groups level. In addition, we assign types to static transforms in launch files which add them to the TF trees, and check for any type errors. After inference, \sysname{} performs additional checks on the inferred types in both source code and launch files which include \emph{sensor\_null}, \emph{reversed\_name} and implicit convention violations checks. 

\section{Threats to Validity}
\label{app:validity}
\noindent
{\em External Threats.} 
The set of benchmarks used in evaluation may not be representative. We mitigate the threat by using a large set of 180 randomly selected projects with various levels of maturity and complexity (with the most complex one having over 7 million LOC). We argue that evaluation on immature projects is equally important as the large group of non-expert ROS developers (of immature projects) are the ones that suffer the most from frame problems. In the future, we foresee to push our type checker to become part of some mainstream robotic software development framework such that the technique can be better evaluated in practice and used during development instead of after deployment.

\smallskip
\noindent
{\em Internal Threats.}
We manually label true positive type errors. Human errors may lead to evaluation bias. 
We mitigate this threat by: 1) reporting some of the findings to the developers and ROS answer forum; 2)
cross-checking the labeling results between two authors; 3) being conservative in labeling true positives.
We mark a case that has any doubt or inconsistent labels (across authors) as a false positive.

\begin{table*}
\captionsetup{justification=centering}
\caption{Inconsistencies and Implicit Convention Violation Warnings \newline {\footnotesize Implicit warnings \#: single value 0 = no warnings were reported for all the thresholds; \newline and value - = a project does not qualify for the implicit convention checks as it does not contain launch files with static transforms. }
}
\label{tab:inconsistencies}
  {\footnotesize
  \begin{tabular}{ p{0.22\textwidth} | p{0.025\textwidth} p{0.03\textwidth} | p{0.075\textwidth} | p{0.04\textwidth} ||
  p{0.23\textwidth} | p{0.025\textwidth} p{0.03\textwidth} | p{0.075\textwidth} | p{0.04\textwidth}}
    \toprule
    \multirow[t]{2}{=}{\rr Github Repository}
    & \multicolumn{2}{p{0.055\textwidth}|}{\rr Inconsistencies}
    & \multirow[t]{2}{=}{\rr Implicit Warnings for z=1,2,5,10 (\#)}
    & \multirow[t]{2}{=}{\rr Run-time (s)} 
    & \multirow[t]{2}{=}{\rr Github Repository}
    & \multicolumn{2}{p{0.055\textwidth}|}{\rr Inconsistencies}
    & \multirow[t]{2}{=}{\rr Implicit Warnings for z=1,2,5,10 (\#)}
    & \multirow[t]{2}{=}{\rr Run-time (s)} \\
    \cmidrule(l){2-3} \cmidrule(l){7-8}
    & TP (\#) & FP (\#) & & 
    & & TP (\#) & FP (\#) & & \\
    \midrule
    pheeno\_ros & 1 & 0 & - & 0 & 
    victim\_localization & 1 & 1 & 0 & 4 \tn
    mavros & 3 & 0 & 0 & 52 & 
    LAM\_SLAM & 1 & 0 & 0 & 62 \tn
    rosflight & 3 & 0 & - & 2 & 
    ipc\_slam\_cnn & 1 & 1 & 1, 1, 0, 0 & 1 \tn
    ur\_modern\_driver & 1 & 0 & - & 13 & 
    denso\_tomato & 0 & 4 & 0 & 5 \tn
    orb\_slam\_2\_ros & 1 & 0 & - & 291 & 
    mbzirc\_task2 & 2 & 0 & 0 & 4 \tn
    gps\_umd & 1 & 0 & - & 1 &
    SpCoSLAM\_Lets & 0 & 1 & 0 & 135 \tn
    \cmidrule(l){1-5}
    se306p1 & 4 & 0 & - & 1 & 
    catkin\_ws & 3 & 2 & 0 & 8 \tn
    roman-technologies & 4 & 0 & - & 6 & 
    robotx & 2 & 0 & 0 & 9 \tn
    CRF & 1 & 0 & 0 & 8 & 
    IntelligentRobotics & 1 & 0 & 0 & 8 \tn
    roscorobot & 2 & 0 & - & 20 & 
    riptide2017 & 1 & 0 & 0 & 0 \tn
    rotors\_simulator\_with\_drcsim\_env & 2 & 0 & - & 6 & 
    air\_manipulator & 12 & 0 & 0 & 23 \tn
    px4\_offboard\_velocity\_control & 1 & 0 & - & 0 & 
    summit-xl-ros-stack & 2 & 0 & 0 & 10 \tn
    velocity\_marker\_tracker & 2 & 0 & - & 0 & 
    pick\_n\_place & 0 & 0 & 2, 2, 0, 0 & 200 \tn
    MultiRobotExplorationPackages & 4 & 1 & 0 & 424 & 
    Mobile-Robotics-Development-Platform & 1 & 0 & 0 & 25 \tn
    body\_axis\_controller & 1 & 0 & - & 0 & 
    jaguar-bot & 2 & 2 & 0 & 53 \tn
    body\_axis\_velocity\_controller & 1 & 0 & - & 0 & 
    semantic\_labels\_sys & 0 & 0 & 1, 0, 0, 0 & 45 \tn
    marker\_tracker & 1 & 0 & - & 1 & 
    Racecar-Platform & 1 & 1 & 0 & 4 \tn
    Loam\_Matlab\_SMP & 1 & 0 & - & 5 &
    fyp\_uav\_radar & 1 & 0 & 0 & 7 \tn
    arduino\_brushless & 1 & 0 & - & 1 & 
    PX4\_Drone & 3 & 0 & 0 & 0 \tn
    frobit\_lego\_transporter & 1 & 0 & - & 1 & 
    depth\_tools & 1 & 0 & 0 & 0 \tn
    wiic\_twist & 1 & 0 & - & 1 & 
    IntelligentCar & 0 & 3 & 0 & 39 \tn
    Create\_PointCloud-ROS & 1 & 0 & - & 1 & 
    rapid\_pbd & 0 & 1 & 0 & 9 \tn
    autorally & 2 & 0 & - & 9 & 
    Lidar\_Utility & 1 & 0 & 0 & 18 \tn
    ua-ros-pkg & 2 & 0 & 0 & 162 & 
    ros\_webcam\_lidar & 1 & 0 & 0 & 0 \tn
    model\_car & 3 & 0 & 0 & 41 & 
    lidar\_slam\_3d & 0 & 1 & 0 & 1 \tn
    door\_detection\_smr & 3 & 0 & - & 4 & 
    Autonomous\_wheelchair\_navigation & 3 & 0 & 0 & 0 \tn
    table\_cleaner\_ur10 & 6 & 0 & - & 18 &  AutoNavQuad & 1 & 0 & 0 & 3 \tn
    \cmidrule(l){1-5}
    ece6460\_examples & 1 & 0 & 0 & 1 & 
    Mini-Lab & 2 & 1 & 0 & 14 \tn
    lsd\_slam & 1 & 0 & 0 & 133 & 
    gamesh\_bridge & 0 & 1 & 0 & 0 \tn
    los\_slam & 1 & 0 & 0 & 2 & 
    rockin\_logger & 1 & 0 & 0 & 1 \tn
    Robotic & 2 & 0 & 0 & 3 & 
    hpp\_source\_code & 0 & 0 & 1, 1, 0, 0 & 9 \tn
    USV-ROS-Stack & 1 & 2 & 0 & 7 & 
    chairbot\_selfdrive & 0 & 0 & 1, 1, 1, 0 & 2 \tn
    Inverse-Kinematics & 2 & 0 & 0 & 2 & 
    mycomputer & 1 & 0 & 0 & 2 \tn
    ifacWC2020 & 0 & 0 & 3, 2, 0, 0 & 17 & 
    NWPU\_robot & 4 & 0 & 0 & 35 \tn
    gki\_pr2\_symbolic\_planning & 0 & 1 & 0 & 7 & match\_catch & 5 & 0 & 2, 2, 2, 0 & 1190 \tn
    util & 2 & 0 & 0 & 1 & 
    HislandAutonomousRobot & 1 & 0 & 0 & 1 \tn
    derail-fetchit-public & 2 & 1 & 2, 1, 1, 0 & 13 & 
    epuck\_world & 1 & 0 & 0 & 0 \tn
    WaterGun2016 & 1 & 0 & 0 & 139 & 
    human\_dialogue & 0 & 2 & 0 & 22 \tn
    PiObs & 2 & 0 & 0 & 0 & 
    monoodometer & 3 & 0 & 0 & 49 \tn
    16-662\_Robot\_Autonomy\_Project & 1 & 0 & 0 & 0 & 
    pushing\_benchmark & 0 & 0 & 5, 5, 2, 0 & 217 \tn
    dyros\_jet\_dg & 1 & 0 & 0 & 69 & 
    hrp\_mpc\_lidar & 1 & 0 & 4, 4, 2, 2 & 868 \tn
    robo-games & 2 & 0 & 0 & 2 & 
    scan\_tools & 0 & 0 & 1, 1, 0, 0 & 3 \tn
    ROS-Challenges & 2 & 0 & 0 & 2 & 
    kill\_test & 0 & 1 & 0 & 18 \tn
    RAS-2016 & 3 & 1 & 0 & 10 & 
    thesis-repo & 3 & 0 & 0 & 1 \tn
    cam\_mocap\_calib & 0 & 3 & 0 & 1 & 
    dashgo & 0 & 0 & 6, 6, 0, 0 & 1 \tn
    bigfoot & 0 & 0 & 1, 1, 0, 0 & 2 & 
    iarc-2017 & 4 & 0 & 1, 1, 0, 0 & 21 \tn
    ros\_project & 4 & 0 & 0 & 5 & 
    iarc-mission8 & 1 & 0 & 1, 0, 0, 0 & 26 \tn
    ae-scoot & 0 & 0 & 7, 7, 7, 7 & 40 & 
    grasping\_oru & 0 & 2 & 0 & 13 \tn
    MR2\_ROS & 0 & 2 & 5, 0, 0, 0 & 1 & 
    navigation\_oru-release & 4 & 0 & 0 & 457 \tn
    Fake-LaserScan & 0 & 0 & 1, 1, 1, 0 & 0 & 
    \tn
    \cmidrule(l){6-10}
    Distributed\_Collaborative\_Task\_Tree\_ubuntu-version-16.04 & 0 & 1 & 0 & 14 & 
    Total & 154 [81.05\%] & 36 & 45, 36, 16, 9 & 5162 \tn
    \bottomrule
  \end{tabular}}
\end{table*}

\section{Case Studies}
\label{app:cases}

\nib{Case I: Redundant transform.}
\begin{figure}[!htbp]
    \centering
    \includegraphics[width=0.95\columnwidth]{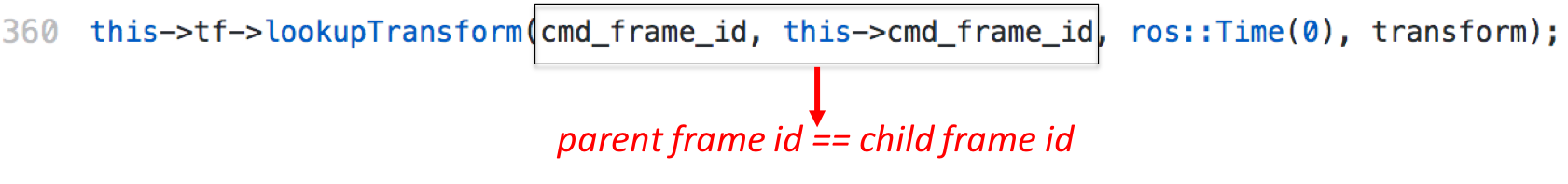}
    \caption{Example for redundant transform {\tiny (source: 
    https://git.io/Jt5Vs
    )}}
    \label{fig:redundant}
\end{figure}
Figure~\ref{fig:redundant} presents a redundant transform detected by \sysname{}. The code invokes the \emph{lookupTransform} function to obtain the transform between identical parent and child frames (`base\_link') from the \emph{tf} tree, which is essentially an identity transform with zero displacements and zero rotation.
Manual analysis found that the transform is used in a a number of places computing distance between the robot (`base\_link') and the target goal. They simply correspond to useless computation. After we reported it to the developer, it was acknowledged that though it does not affect the correctness, it is completely unnecessary and can cause confusion.

\nib{Case II: Missing frame.}
\begin{figure}[!htbp]
    \centering
    \includegraphics[width=0.95\columnwidth]{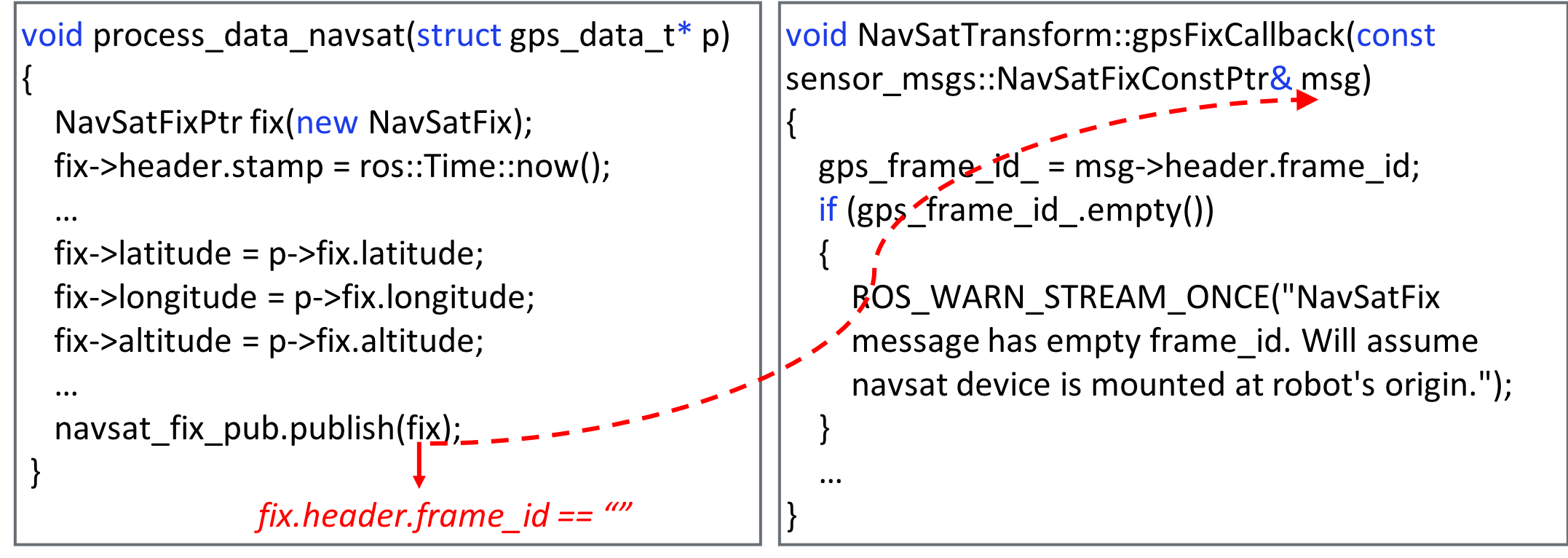}
    \caption{Example for missing frame information in the published ROS message {\tiny (fixed in: https://git.io/JtSJv
    )}}
    \label{fig:missing}
\end{figure}
Figure~\ref{fig:missing} shows the code snippet related to a missing frame error reported by \sysname. It was also reported in \emph{swri-robotics/gps\_umd} GitHub repository. The function on the left is part of this repository and is responsible for publishing navigation satellite (i.e., GPS) fix, whereas the function on the right is from another package (or repository) called \emph{robot\_localization} and uses the published fix data to fuse into robot's position estimate.  Since, \emph{robot\_localization} transforms GPS data into a frame that matches robot’s starting pose (position and orientation) in its world frame, it needs the frame information of the received GPS data. However, the function on the left does not specify the frame information in the published fix, and hence \emph{robot\_localization} assumes that the navsat device is mounted at robot's origin. This can be problematic if the navsat device is mounted at different pose. This illustrates that the frame information of the published data becomes important during integration of different packages. Moreover, the missing frame inconsistency makes code harder to understand and maintain. 

\nib{Case III: Cycle in the frame tree:}
\begin{figure}[!htbp]
    \centering
    \includegraphics[width=\columnwidth]{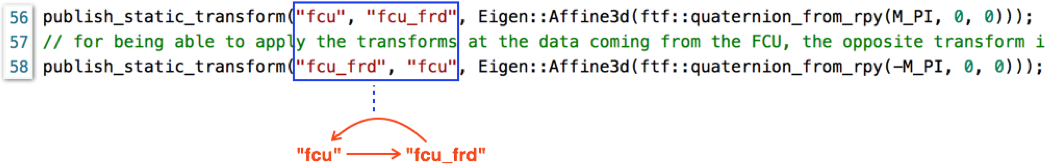}
    \caption{Example of loop in the tf tree {\tiny (fixed in: https://git.io/Jt5VB)}}
    \label{fig:cycle}
\end{figure}
Figure~\ref{fig:cycle} illustrates an example of a loop in the TF tree. Observe that the two published transforms {\it fcu}$\rightarrow${\it fcu\_frd} and {\it fcu\_frd}$\rightarrow${\it fcu} form a cycle. The \emph{tf} package maintains parent-child relationships between the frames in a directed tree structure which by definition cannot have cycles. A cycle invalidates a tree structure definition as well as can lead to multiple paths between two nodes resulting in more than one net transforms between the corresponding frames, which in turn causes ambiguity during the computation of a net transform between two frames.

\begin{table*}
\captionsetup{justification=centering}
\caption{Projects used in the evaluation of \sysname{} \newline {\footnotesize C=C/C++ source files, L=Launch files, LOC=non-blank and non-commented lines of C/C++ source and header files as computed by CLOC}}
\label{tab:projects}
  {\footnotesize
  \begin{tabular}{ p{0.18\textwidth}|p{0.02\textwidth} p{0.02\textwidth} p{0.03\textwidth}||p{0.18\textwidth}|p{0.02\textwidth} p{0.02\textwidth} p{0.03\textwidth}||p{0.18\textwidth}|p{0.02\textwidth} p{0.02\textwidth} p{0.03\textwidth}}
  \toprule
  Github Repository & C(\#) & L(\#) & LOC & Github Repository & C(\#) & L(\#) & LOC & Github Repository & C(\#) & L(\#) & LOC\\
  \midrule
cam\_mocap\_calib & 2 & 3 & 397 &
bigfoot & 10 & 14 & 1282 &
ros\_project & 16 & 4 & 3224 \tn
ae-scoot & 44 & 15 & 12405 &
jaguar-bot & 96 & 52 & 20089 &
tbot\_arm\_ws & 1 & 2 & 65 \tn
nbvplanner & 11 & 10 & 3291 &
MR2\_ROS & 5 & 6 & 1459 &
depth\_tools & 4 & 2 & 473 \tn
IntelligentCar & 86 & 44 & 46421 &
semantic\_labels\_sys & 102 & 151 & 21031 &
LAM\_SLAM & 151 & 30 & 25638 \tn
ipc\_slam\_cnn & 3 & 3 & 2769 &
rapid\_pbd & 33 & 15 & 7361 &
Baxter-Uncertainty & 3 & 3 & 3783\tn
DaVinci-Haptics & 1 & 3 & 314 &
Racecar-Platform & 10 & 22 & 3885 &
PX4\_Drone & 3 & 3 & 236 \tn
A2DR\_CHULA & 21 & 37 & 9221 &
csir-manipulation & 3 & 9 & 465 &
robotx & 25 & 7 & 5698 \tn
robond\_project\_6 & 4 & 5 & 71 &
IntelligentRobotics & 17 & 37 & 7435 &
ME495Final & 1 & 6 & 229 \tn
velodyne\_driver & 17 & 7 & 1799 &
Robotics & 28 & 30 & 4814 &
crazy\_lidar & 5 & 3 & 1698 \tn
crazypi & 1 & 2 & 471 &
RAS-2016 & 43 & 34 & 7835 &
denso\_tomato & 20 & 12 & 4458 \tn
mbzirc\_task2 & 17 & 29 & 3422 &
ros\_assignments & 15 & 2 & 926 &
catkin\_ws & 44 & 66 & 5872 \tn
Robotics-Localization & 2 & 6 & 28 &
team2 & 7 & 17 & 18263 &
team3 & 8 & 10 & 18499 \tn
AGV\_ws & 16 & 31 & 6799 &
16-662\_Robot\_Autonomy\_Project & 3 & 5 & 103 &
dyros\_jet\_dg & 34 & 24 & 16401 \tn
kobuki\_navigation & 1 & 10 & 163 &
quadrotor\_moveit\_nav & 5 & 13 & 336 &
robo-games & 6 & 17 & 1124 \tn
ROS-Challenges & 8 & 16 & 1728 &
mbot\_arm\_DRL & 1 & 3 & 103 &
ArDroneControl & 6 & 7 & 1585 \tn
kuka\_lwr\_simulation & 10 & 59 & 2177 &
contamination\_stack & 5 & 6 & 47856 &
VehicleControl & 147 & 38 & 31301 \tn
riptide2017 & 2 & 6 & 507 &
air\_manipulator & 78 & 27 & 13507 &
udacity\_bot & 1 & 3 & 28 \tn
summit-xl-ros-stack & 28 & 187 & 173054 &
fyp\_uav\_radar & 33 & 64 & 445617 &
ece6460\_examples & 29 & 14 & 1486 \tn
lsd\_slam & 48 & 12 & 16489 &
at\_home\_rsbb & 14 & 6 & 3873 &
los\_slam & 8 & 5 & 1434 \tn
Robotic & 18 & 17 & 2164 &
udacity\_bot-d & 1 & 3 & 29 &
USV-ROS-Stack & 27 & 16 & 5323 \tn
Inverse-Kinematics & 18 & 20 & 2380 &
ifacWC2020 & 30 & 8 & 9102 &
Udacity-RoboND-project-2-Kinematics-Project-master & 3 & 18 & 1199 \tn
gki\_pr2\_symbolic\_planning & 49 & 3 & 7461 &
robot\_2d\_nav & 3 & 1 & 362 &
NASA-RMC-2020-NorthstarRobotics & 21 & 37 & 2436 \tn
util & 4 & 4 & 970 &
carl\_bot & 14 & 17 & 2745 &
carl\_moveit & 2 & 9 & 1118 \tn
derail-fetchit-public & 39 & 29 & 10420 &
rail\_ceiling & 3 & 11 & 846 &
WaterGun2016 & 93 & 14 & 15151 \tn
PiObs & 1 & 5 & 17 &
Where-Am-I-Udacity & 2 & 6 & 29 &
victim\_localization & 16 & 3 & 2667 \tn
Fake-LaserScan & 1 & 1 & 44 &
pick\_n\_place & 37 & 6 & 21281 &
Mobile-Robotics-Development-Platform & 74 & 21 & 14200 \tn
RosBat & 6 & 7 & 1261 &
ultrasonic\_to\_laserscan & 5 & 1 & 532 &
gps\_umd & 3 & 0 & 452 \tn
mavros & 69 & 13 & 13065 &
orb\_slam\_2\_ros & 59 & 5 & 22460 &
pheeno\_ros & 5 & 4 & 549 \tn
realsense-ros & 3 & 12 & 1771 &
rosflight & 9 & 0 & 1850 &
ur\_modern\_driver & 22 & 12 & 5697 \tn
door\_detection\_smr & 17 & 3 & 8127 &
table\_cleaner\_ur10 & 62 & 50 & 17235 &
ua-ros-pkg & 125 & 123 & 38418 \tn
model\_car & 22 & 16 & 13928 &
autorally & 35 & 42 & 9220 &
arduino\_brushless & 4 & 0 & 447 \tn
frobit\_lego\_transporter & 2 & 5 & 195 &
RoboticBugAlgo & 8 & 0 & 1170 &
wiic\_twist & 3 & 0 & 795 \tn
Create\_PointCloud-ROS & 6 & 0 & 384 &
CS491\_Autonomous\_Buggy & 74 & 15 & 28874 &
MultiRobotExplorationPackages & 206 & 78 & 7115053 \tn
body\_axis\_controller & 1 & 0 & 61 &
body\_axis\_velocity\_controller & 1 & 0 & 59 &
marker\_tracker & 1 & 3 & 76 \tn
px4\_offboard\_velocity\_control & 1 & 0 & 214 &
rotors\_simulator\_with\_drcsim\_env & 27 & 47 & 4579 &
velocity\_marker\_tracker & 1 & 0 & 53 \tn
Loam\_Matlab\_SMP & 6 & 3 & 2209 &
roscorobot & 47 & 19 & 14831 &
CRF & 23 & 4 & 3386 \tn
roman-technologies & 23 & 0 & 13102 &
catkin\_roboway & 46 & 38 & 25360 &
NWPU\_robot & 95 & 59 & 23834 \tn
baxter\_project & 413 & 20 & 79454 &
AprilTag\_Localization & 93 & 22 & 196152 &
autonomus\_transport\_industrial\_system & 2 & 5 & 561 \tn
gx2019\_omni\_simulations & 5 & 12 & 921 &
uwb-localization & 23 & 3 & 6324 &
match\_catch & 65 & 146 & 135516 \tn
Lidar\_Utility & 72 & 41 & 9300 &
ros\_leishen\_lidar & 2 & 1 & 253 &
ros\_webcam\_lidar & 3 & 2 & 253 \tn
pose\_estimation & 3 & 3 & 1083 &
lidar\_slam\_3d & 4 & 1 & 885 &
surgical\_robot & 1 & 12 & 17 \tn
Autonomous\_wheelchair\_navigation & 1 & 21 & 94 &
slam\_autonomous\_navigation & 38 & 14 & 8450 &
F1tenth & 5 & 11 & 171 \tn
AutoNavQuad & 17 & 26 & 3492 &
SpCoSLAM\_Lets & 63 & 11 & 11488 &
Mini-Lab & 49 & 66 & 6941 \tn
gamesh\_bridge & 1 & 2 & 74 &
rockin\_logger & 1 & 2 & 160 &
head\_meka & 0 & 1 & 0 \tn
meka\_pmb2\_ros & 4 & 59 & 398 &
hpp\_source\_code & 0 & 2 & 0 &
RoboND-Localization-Project & 2 & 6 & 29 \tn
amcl\_3d & 11 & 2 & 1518 &
autoware\_rc\_car & 5 & 6 & 57767 &
rc\_car & 3 & 9 & 200 \tn
vehicle\_sim & 2 & 7 & 196 &
ros\_lecture19\_pkg & 5 & 1 & 179 &
keti\_ws & 76 & 13 & 42784 \tn
ros-detection & 1 & 3 & 12822 &
chairbot\_selfdrive & 10 & 20 & 1497 &
tms\_ss\_pot & 12 & 10 & 2565 \tn
mycomputer & 14 & 50 & 5079 &
epuck\_world & 2 & 3 & 531 &
Distributed\_Collaborative\_Task\_Tree\_ubuntu-version-16.04 & 46 & 73 & 8867 \tn
human\_dialogue & 60 & 68 & 13760 &
ENPM-661-Planning-Projects & 1 & 1 & 67 &
monoodometer & 73 & 16 & 17056 \tn
3D-Mapping-Navigation-of-Quadrotors & 2 & 17 & 2015 &
pushing\_benchmark & 58 & 25 & 62266 &
tangentbug & 2 & 1 & 332 \tn
hrp\_mpc\_lidar & 154 & 95 & 108073 &
pioneer\_tools & 5 & 10 & 230 &
turtlesim\_pioneer & 5 & 1 & 266 \tn
rcah18\_pepper\_navigation & 1 & 11 & 280 &
se306p1 & 16 & 0 & 1851 &
sequential\_scene\_parsing & 22 & 4 & 8755 \tn
3d\_mapping & 6 & 21 & 629 &
scan\_tools & 12 & 8 & 2346 &
ROSBeginner & 4 & 10 & 125 \tn
px4VIO & 9 & 1 & 1491 &
ua-catvehicle-dev & 5 & 24 & 567 &
octagon & 24 & 14 & 2425 \tn
kill\_test & 50 & 26 & 15288 &
contactdb\_utils & 11 & 12 & 1620 &
thesis-repo & 33 & 5 & 860 \tn
oko\_slam & 29 & 23 & 7381 &
dashgo & 4 & 19 & 3346 &
HislandAutonomousRobot & 6 & 6 & 632 \tn
iarc-2017 & 45 & 56 & 10578 &
iarc-mission8 & 27 & 39 & 14238 &
theLearner & 11 & 1 & 1290 \tn
onigiri\_war & 1 & 8 & 106 &
Mini-Lab-8 & 8 & 39 & 626 &
grasping\_oru & 15 & 16 & 6911 \tn
navigation\_oru-release & 134 & 95 & 84833 &
yumi\_demos & 3 & 12 & 917 &
orsens\_ros & 2 & 4 & 1137 \tn
  \bottomrule
  \end{tabular}}
\end{table*}
\fi

\end{document}